\documentclass{article}

\PassOptionsToPackage{numbers}{natbib}

\usepackage[final]{neurips_2024}

\usepackage[utf8]{inputenc} 
\usepackage[T1]{fontenc}    
\usepackage{hyperref}       
\usepackage{url}            
\usepackage{booktabs}       
\usepackage{amsfonts}       
\usepackage{nicefrac}       
\usepackage{microtype}      
\usepackage{xcolor}         
\usepackage{comment}

\usepackage{caption}
\usepackage{subcaption}
\usepackage{graphicx}
\usepackage{multirow}
\usepackage{tabularx}
\usepackage{amsmath}
\usepackage{amssymb}
\usepackage{algorithmic}
\usepackage{algorithm}
\usepackage{wrapfig}
\usepackage{rotating}
\usepackage{color, colortbl}
\definecolor{LightCyan}{rgb}{0.88,1,1}
\definecolor{LightGray}{rgb}{0.93,0.93,0.93}

\title{Introducing Spectral Attention for \\Long-Range Dependency in Time Series Forecasting}

\author{%
  Bong Gyun Kang$^{1}$\thanks{Denotes equal contribution. \{luckypanda, elite1717\}@snu.ac.kr} \\
  \And
  Dongjun Lee$^{1*}$\\
  \And
  HyunGi Kim$^{2}$\\
  \And
  DoHyun Chung$^{3}$\\
  \And
  Sungroh Yoon$^{1,2}$\thanks{Corresponding author.  sryoon@snu.ac.kr}
  \AND
  {\normalfont$^1$ Interdisciplinary Program in Artificial Intelligence, Seoul National University}\\
  $^2$ Department of Electrical and Computer Engineering, Seoul National University\\
  $^3$ Department of Future Automotive Mobility, Seoul National University\\
}

\begin{document}

\maketitle

\begin{abstract}
Sequence modeling faces challenges in capturing long-range dependencies across diverse tasks. Recent linear and transformer-based forecasters have shown superior performance in time series forecasting. 
However, they are constrained by their inherent inability to effectively address long-range dependencies in time series data, primarily due to using fixed-size inputs for prediction. 
Furthermore, they typically sacrifice essential temporal correlation among consecutive training samples by shuffling them into mini-batches.
To overcome these limitations, we introduce a fast and effective Spectral Attention mechanism, which preserves temporal correlations among samples and facilitates the handling of long-range information while maintaining the base model structure.
Spectral Attention preserves long-period trends through a low-pass filter and facilitates gradient to flow between samples.
Spectral Attention can be seamlessly integrated into most sequence models, allowing models with fixed-sized look-back windows to capture long-range dependencies over thousands of steps.
Through extensive experiments on 11 real-world time series datasets using 7 recent forecasting models, we consistently demonstrate the efficacy of our Spectral Attention mechanism, achieving state-of-the-art results.
\end{abstract}

\section{Introduction}
Time series forecasting (TSF) stands as a core task in machine learning, ubiquitous in our lives through applications such as weather forecasting, traffic flow estimation, and financial investment. Over time, TSF techniques have evolved from statistical~\cite{box1968some, GaussianProcess, hyndman2018forecasting} and machine learning approaches~\cite{ahmed2010empirical, friedman2001greedy, SVM} to deep learning models like Recurrent Neural Networks~\cite{hochreiter1997long, lai2018modeling, sagheer2019time} and Convolution based Networks~\cite{hewage2020temporal, wan2019multivariate}.
Following the success of Transformers~\cite{Transformer} in various domains, Transformer-based models have also become mainstream in the time series domain~\cite{Reformer, LogSparseTransformer, PatchTST, Autoformer, Crossformer, Informer, Fedformer}. Recently, methodologies based on Multi-layer Perceptron have received renewed attention~\cite{TiDE, TSMixer, RLinear, FreTS}. 
However, despite the advancements, long-term dependency modeling in TSF remains challenging~\cite{DLinear}.

\begin{figure}[t]
    \vspace{-15pt}
    \centering
    \begin{subfigure}[b]{1.0\columnwidth}
        \includegraphics[width=\linewidth]{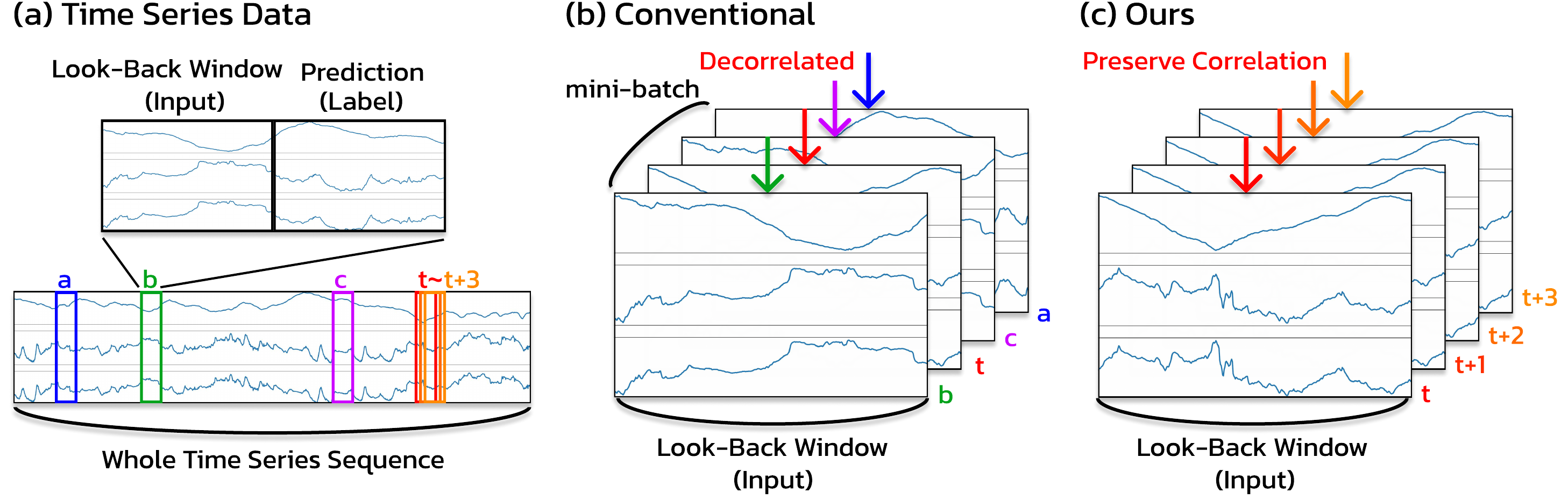}
    \end{subfigure}
    \vspace{-15pt}
    \caption{(a) Training data are sampled for each time step from continuous sequences, exhibiting high temporal correlations. (b) Conventional approaches simply ignore this temporal information with a shuffled batch. (c) We address the temporal correlation between the samples for the first time, enabling the model to consider long-range dependencies that surpass the look-back window.}
    \label{fig:intro}
    \vspace{-14pt}
\end{figure}

Unlike image models, where data are randomly sampled from the image distribution and are thus independent of each other~\cite{hsu2019measuring}, TSF models sample data from the continuous signal, dependent on the time variable $t$ as shown in Figure~\ref{fig:intro}a. 
This leads to a high level of correlation between each training sample, which consists of a fixed-sized look-back window before $t$ (as input) and the subsequent prediction sequence after $t$ (as label).
Therefore, the conventional approach of shuffling the consecutive samples into mini-batches deprives the model of utilizing the crucial inherent temporal correlation between the samples. This restricts the model’s consideration to only a fixed-size look-back window for sequence modeling, limiting the ability to address long-range dependencies (Figure~\ref{fig:intro}b).


Recent studies pointed out that simply increasing the look-back window leads to substantially detrimental effects such as increased model size and longer training and inference times~\cite{DLinear}. This is particularly challenging for transformer forecasters, which exhibit quadratic time/memory complexity~\cite{PatchTST, Transformer, Crossformer}, and even for efficient models using techniques like Sparse Attention, which have O(nlogn) complexity~\cite{Reformer, LogSparseTransformer, Informer}. Furthermore, if the commonly used look-back window of 96 is extended fivefold, the model can only utilize time steps of less than 500, making it difficult to consider long-range dependencies spanning thousands or the entire dataset. Also, increasing the look-back window may not be beneficial, often leading to decreased performance~\cite{DLinear}, highlighting the fact that current models are not sufficient in capturing long-range dependencies. 

To address this limitation, we propose Spectral Attention, which can be applied to most TSF models and enables the model to utilize long-range temporal correlations in sequentially obtained training data. With the stream of consecutive training samples (Figure~\ref{fig:intro}c), Spectral Attention stores an exponential moving average (EMA) of the activations with various smoothing factors at each time step. This serves as a low-pass filter, inherently embedding long-range information over a thousand steps. Attention is then applied to the stored EMA activations of various smoothing factors (low-pass filters with different frequencies), enabling the model to learn which periodic trends to consider when predicting the future, thereby enhancing its performance. Spectral Attention is even applicable to models such as iTransformer~\cite{iTransformer}, which do not preserve the temporal order of time series data internally.

We further extend Spectral Attention, where computations depend on the activation of the previous timesteps, to Batched Spectral Attention, enabling parallel training across multiple timesteps.
This extension makes Spectral Attention faster and more practical and allows for the direct utilization of temporal relationships among consecutive data samples within a mini-batch in the training base model. In Batched Spectral Attention, the EMA is unfolded over time to perform Spectral Attention simultaneously across multiple time steps. This unfolding allows gradients at time $t$ to propagate through the Spectral Attention module to the previous time step within the mini-batch, achieving effects similar to Backpropagation Through Time (BPTT)~\cite{bptt} and extends the model's effective input window.

\textbf{Our approach} preserves the base TSF model architecture and learning objective while enabling the model to leverage long-term trends spanning thousands of steps. By effectively utilizing the temporal correlation of training samples, our method allows gradients to flow back in time beyond the look-back window, extending to the entire mini-batch. Also, our method requires little additional training time and memory. We conducted experiments on 7 recent TSF models and 11 real-world datasets and demonstrated consistent performance enhancement in all architecture, achieving state-of-the-art results. 
We summarize the main contributions as follows:
\begin{itemize}
\item We propose Spectral Attention, which addresses long-range dependencies spanning thousands of steps through frequency filtering and attention mechanisms, leveraging the temporal correlation among the consecutive samples.
\item We propose Batched Spectral Attention, which enables parallel training across multiple timesteps and expends the effective input window, allowing the gradient to flow through time within the mini-batch.
\item Batched Spectral Attention is applicable to most existing TSF models and practical in real-world scenarios with minimal additional memory and comparable training time. Also, it allows finetuning with a trained TSF model.
\item Batched Spectral Attention demonstrates consistent model-agnostic performance improvements, particularly showcasing superior performance on datasets with significant long-term trend variations.
\end{itemize}

\section{Related Works}
\textbf{Classic TSF models.}
Statistical TSF methods, such as ARIMA~\cite{arima}, Holt-Winters~\cite{hyndman2018forecasting}, and Gaussian Process~\cite{GaussianProcess}, assume that temporal variations adhere to predefined patterns. However, their practical applicability is largely limited by the complex nature of real-world data. Machine learning approaches, such as Support Vector Machines~\cite{boser1992training} and Random Forests~\cite{ho1995random} have proven to be effective even compared to early artificial neural networks~\cite{hinton2006fast, hu1964adaptive, rosenblatt1957perceptron}.
%
Convolutional network-based methods leverage convolution kernels to capture temporal variations sliding along the temporal dimension~\cite{hewage2020temporal, wan2019multivariate}. Recurrent neural network (RNN) grasp changes over time via state transitions across different time steps.~\cite{lai2018modeling, sagheer2019time}. However, RNN-based models exhibit limitations in modeling long-range dependencies due to challenges such as vanishing gradients and error accumulation~\cite{liu2020dstp, sutskever2014sequence}. 
Recently, transformer and linear-based models have emerged as alternatives, demonstrating superior performance compared to recurrent models~\cite{wen2022transformers, DLinear}.

\textbf{Transformer and Linear based models.}
Transformer-based models~\cite{Transformer} address temporal relationships between time points using the attention. LogSparseTransformer~\cite{LogSparseTransformer}, Reformer~\cite{Reformer}, and Informer~\cite{Informer} have been proposed to make the Transformer architecture efficient, addressing the quadratic time complexity. 
The Autoformer~\cite{Autoformer} incorporates series decomposition as an inner block of Transformer and aggregates similar sub-series by utilizing the Auto-Correlation mechanism. 
PatchTST~\cite{PatchTST} introduces patching, a channel-independent approach that processes each variable separately and focuses on cross-time attention. Crossformer~\cite{Crossformer} utilizes a channel-dependent approach to learn cross-variate dependencies. This is achieved through the use of cross-time and cross-dimension attention.
iTransformer~\cite{iTransformer} applies attention and FFN in an inverted way, where attention handles correlations between channels and FFN handles the temporal information. 
%
Recently, to address Transformers' potential difficulties in capturing long-range dependencies~\cite{DLinear}, methodologies based on the linear model and Multi-Layer Perceptron (MLP) structures have emerged. 
DLinear~\cite{DLinear} utilizes the decomposition method introduced in Autoformer and predicts by adding the output of two linear layers for each seasonal and trend element.
TiDE~\cite{TiDE} proposes an architecture based on MLP residual blocks that combines information from dynamic and static covariates with a look-back window for encoding, followed by decoding. 
TSMixer~\cite{TSMixer} performs forecasting by repeatedly mixing time and features using an MLP. 
RLinear~\cite{RLinear} comprises of a single linear layer with RevIN~\cite{RevIN} for normalization and de-normalization.

\textbf{Frequency-utilizing models. }
Using the frequency domain for TSF is a well-established approach~\cite{bloomfield2004fourier, 9446858, 9770652}.
Conventional approaches leverage frequency information during the preprocessing stage~\cite{pandiyan2020analysis} or decompose time series based on frequency filtering~\cite{rezaei2021stock}.
In deep TSF models, research has also explored architectural advancements that are aware of the frequency information. 
SAAM~\cite{moreno2023deep}, which is applicable to RNNs, performs FFT and autocorrelation on the input signal.
WaveFroM~\cite{WaveForM} uses discrete wavelet transform (DWT) to project time series into wavelet domains of various scales and performs forecasting through graph convolution and dilated convolution. 
FEDformer~\cite{Fedformer} adopts a mixture-of-experts strategy to refine the decomposition of seasonal and trend components and introduce sparse attention mechanisms in the frequency domain. 
TimesNet~\cite{Timesnet} transforms 1D time series into 2D tensors utilizing multi-periodicity by identifying dominant frequencies through Fourier Transform, modeling temporal variations effectively.
FreTS~\cite{FreTS} leverages frequency-domain MLPs to achieve global signal analysis and compact energy representation, addressing the limitations of point-wise mappings and information bottlenecks in conventional MLP-based methods. 
FITS~\cite{FITS} employs interpolation within the complex frequency domain to construct a concise and robust model.
While these models leverage frequency information, they are limited in modeling long-range dependencies, as the frequency conversion is confined to the look-back window. 
On the other hand, our Spectral Attention is the first to achieve long-range dependency modeling beyond the look-back window by incorporating consecutive data streams during model training.

\section{Methods}
\begin{figure}[t]
    \vspace{-15pt}
    \centering
    \begin{subfigure}[b]{1.\columnwidth}
        \includegraphics[width=\linewidth]{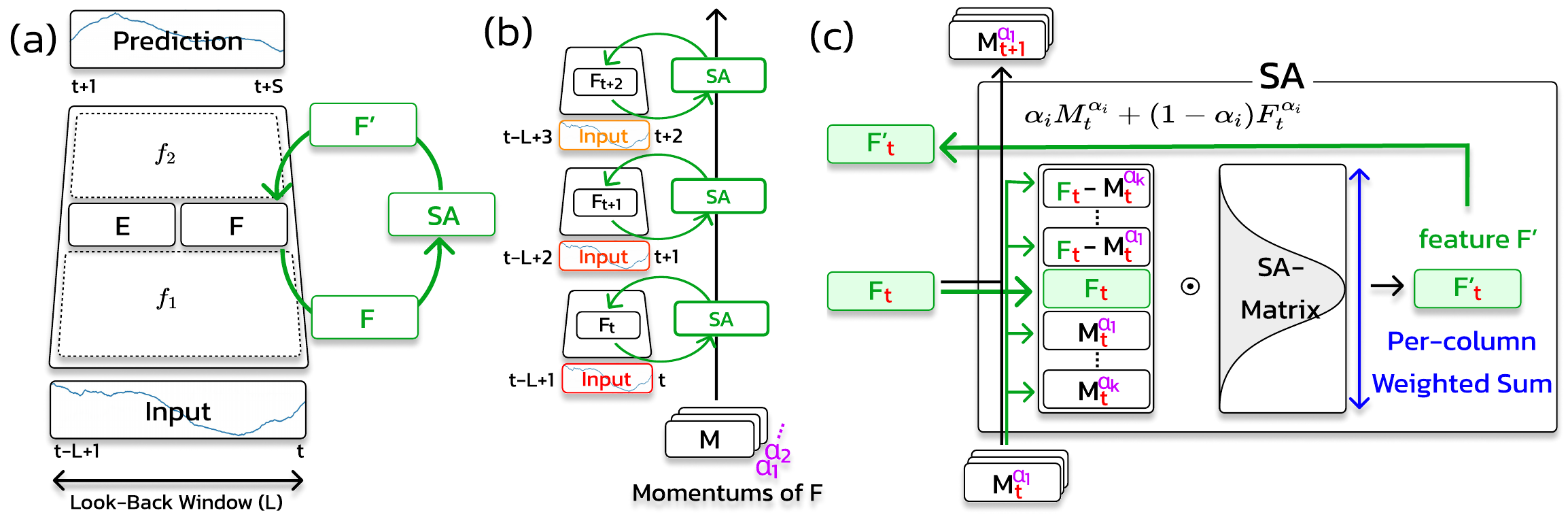}
    \end{subfigure}
    \vspace{-10pt}
    \caption{
    (a) Plug-in Spectral Attention (SA) module takes a subset of intermediate feature $F$ and returns $F'$ with long-range information beyond the look-back window. The model is trained end-to-end, and gradients flow through the SA module. (b) To capture the long-range dependency, SA stores momentums of feature $F$ generated from the sequential inputs. Multiple momentum parameters $\alpha_i$ capture dependencies across various ranges. (c) SA module computes $F'$ by attending multiple low-frequency ($M^{\alpha_i}$) and high-frequency ($F-M^{\alpha_i}$) components and feature ($F$) using learnable Spectral Attention Matrix (SA-Matrix)}
    \label{fig:sa_3}
    \vspace{-10pt}
\end{figure}
\textbf{Problem Statement.} In multivariate time series forecasting, time series data is given $\mathbb{D}_T: \left \{ x_1,..., x_T \right \} \in \mathbb{R}^{T\times N}$ at time $T$ with $N$ variates. Our goal is, at arbitrary future time $t>T$, to predict future $S$ time steps $Y_t=\left \{ x_{t+1}, ...,x_{t+S} \right \}\in \mathbb{R}^{S\times N}$. 
To achieve this goal, TSF model $f$ utilizes length $L$ look-back window as input $X_t=\left \{ x_{t-L+1}, ...,x_{t} \right \}\in \mathbb{R}^{L\times N}$ making prediction $P_t = f(X_t)$, $P\in \mathbb{R}^{S\times N}$.

Model is trained with the training dataset $D_T=\left \{ (X_t, Y_t) | L\leq t\leq T-S \right \}$. While conventional methods typically randomly sample each $X, Y$ from $D_T$ to constitute the mini-batch, we utilize sequential sampling to incorporate temporal correlations between samples into the learning process.

\subsection{Spectral Attention}
Spectral Attention ($SA$) can be applied to every TSF model that satisfies the aforementioned problem statement. This base TSF model is represented by $P=f(X)$, and $SA$ can be applied to arbitrary activation $F$ within the model. The base model can be reformulated as $P=f_2(F, E)$ and $F, E = f_1(X)$. 
$F, E$ are intermediate state and SA module takes an arbitrary subset $F$ as input and transforms it into $F'$ of the same size; $F'=SA(F)$, $P'=f_2(F', E)$. The resulting SA plugged model $f_{SA}$ is depicted in Figure~\ref{fig:sa_3}a.

With $X_t$ as the base model input, $SA$ takes $D$-dimensional feature vector $F_t\in \mathbb{R}^{D}$ as input. $SA$ updates the exponential moving average (EMA) $M_t\in \mathbb{R}^{K \times D}$ of $F_t$ in its internal memory with the $K$ \textit{smoothing factors} $\left \{\alpha_{1},...,\alpha_{K}\right \} \in \mathbb{R}^{K}$ $(\alpha_1<\cdots <\alpha_K)$ as shown in Figure~\ref{fig:sa_3}b.
\begin{gather}
\label{eq:1}
    M^{k,i}_{t+1} = \alpha_{k}\times M^{k,i}_{t} + (1-\alpha_{k})\times F^{i}_t
\end{gather}
EMA retains the trend of features over long-range time periods based on the smoothing factor. It operates as a low-pass filter, with the -3db (half) cut-off frequency of $freq_{cut} = \frac{1}{2\pi }cos^{-1}\left [ 1-\frac{(1-\alpha)^2}{2\alpha} \right ]$, effectively preserving the trend over 6,000 period with $\alpha = 0.999$ .

To represent high-frequency patterns contrasting with the low-pass filtered long-range pattern $M_t$, we generated $H_t\in \mathbb{R}^{K \times D}$ by subtracting $M_t$ from $F_t$.
\begin{gather}
\label{eq:2}
    H^{k,i}_{t} = F^{i}_t - M^{K-k-1,i}_{t}
\end{gather}
$SA$ contains learnable parameters: \textit{sa-matrix} $\in \mathbb{R}^{(2K+1) \times D}$, 
which learns what frequency the model should attend to for each feature. $2 \times H_t$, $F_t$, $2\times M_t$ are concatenated on dimension 0, resulting in $\mathbb{R}^{(2K+1) \times D}$, which is then weighted summed with \textit{sa-matrix} on dimension 0, generating output $F'_t$ (Figure~\ref{fig:sa_3}c).
\begin{gather}
\label{eq:3}
    F'_{t} = sum(softmax(\textrm{\textit{sa-matrix}}, dim\ 0) \cdot concat((2 \times H_t, F_t, 2\times M_t), dim\ 0), dim\ 0)
\end{gather}
The \textit{sa-matrix} is initialized so that $softmax(\textrm{\textit{sa-matrix}})$ resembles a Gaussian distribution on axis 0.
This results in symmetric value on axis 0 (\textit{sa-matrix}$^{K+1-i}$ = \textit{sa-matrix}$^{K+1+i}$) and makes $SA$ an identity function on initialization ($\because  H^k + M^{K-k-1} = F$).
\begin{gather}
\label{eq:4}
    F=SA_{init}(F)
\end{gather}
$SA$ allows the model to attend to multiple frequencies of its feature signal, enabling it to focus on either long-range dependencies or high-frequency patterns as needed and shift the feature $F$ distribution on the frequency domain. 
By initializing $SA$ as an identity function, the model can be fine-tuned with the already trained base model, allowing efficient implementation.

\subsection{Batched Spectral Attention}
\begin{figure}[t]
    \vspace{-15pt}
    \centering
    \begin{subfigure}[b]{1.\columnwidth}
        \includegraphics[width=\linewidth]{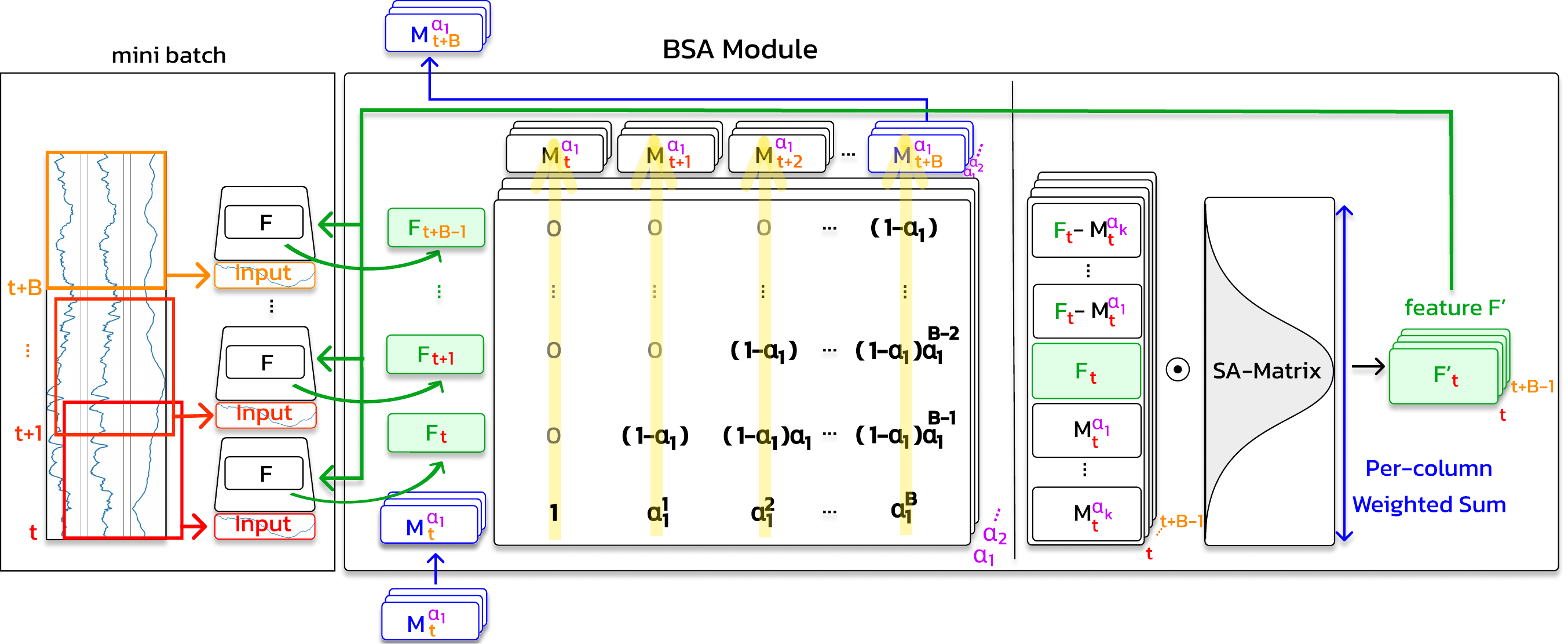}
    \end{subfigure}
    \vspace{-15pt}
    \caption{
    BSA module takes a sequentially-sampled mini batch $ \left\{X_t,...X_{t+B-1} \right\}$ and computes the corresponding EMA momentums  $\left\{M_t,...M_{t+B-1} \right\}$ over time. This is done via single matrix multiplication enabling parallelization. We made the momentum parameter $\alpha_i$ learnable, allowing the model to directly learn the periodicity of the information essential for the future prediction.}
    \label{fig:method}
    \vspace{-10pt}
\end{figure}

Batched Spectral Attention ($BSA$) enables batch training over multiple time steps. The main concept involves unfolding EMA, which facilitates gradients to flow across consecutive samples in a mini-batch, akin to BPTT. This enables efficient parallel training and promotes the model to extract long-range information beneficial for future prediction, extending the effective look-back window. The overall flow of $BSA$ is depicted in Figure~\ref{fig:method}.

With mini-batch of size $B$, consecutive samples $X_{\left [t, t+B-1\right ]} = \left \{ X_{t}, ... X_{t+B-1} \right \} \in \mathbb{R}^{B\times S\times N}$ are given as input. Following aforementioned $SA$ setting, $BSA$ takes $F_{\left [t, t+B-1\right ]} = \left \{ F_{t}, ... F_{t+B-1} \right \}\in \mathbb{R}^{B \times D}$ as input. 
In the next step, $BSA$ utilizes $F_{\left [t, t+B-1\right ]}$ and the stored $M_t\in \mathbb{R}^{K \times D}$ to generate $M_{t+b} (0 \leq b \leq B)$ by unfolding the Equation \ref{eq:1}.
\begin{gather}
\label{eq:5}
    M^{k,i}_{t+b} = \alpha^{b}_{k}\times M^{k,i}_{t} + (1-\alpha_{k})\alpha^{b-1}_{k}\times F^{i}_{t}+\cdots+(1-\alpha_{k})\times F^{i}_{t+b-1}
\end{gather}
This equation can be transformed to calculate $M_{\left [t, t+B\right ]}\in \mathbb{R}^{(B+1) \times K \times D}$ in parallel as follows.
\begin{gather}
\label{eq:6}
    M^{:,k,i}_{\left [t, t+B\right ]} = lower\textrm{-}triangle(A^{k}) \times concat((M^{k,i}_{t}, F^{:,i}_{\left [t, t+B-1\right ]}), dim \ 0)\\
    A \in \mathbb{R}^{K\times (B+1)\times (B+1)}, \ A^{k,p,q} = (1-\alpha_{k} )^{I\left \{  q>0\right \}}\alpha^{p-q}_{k} 
\end{gather}
$A$ refers to unfolding matrix and $I$ refers to indicator function. $M_{t+B}$ is stored in $BSA$ for the next mini-batch input. 
$F'_{\left [t, t+B-1\right ]}$ is computed in parallel, similar to Equation \ref{eq:2} and \ref{eq:3}, using $F_{\left [t, t+B-1\right ]}$, $B_{\left [t, t+B-1\right ]}$, and \textit{sa-matrix}$\in \mathbb{R}^{(2K+1) \times D}$.
The $lower\textrm{-}triangle$ function prevents gradients from the past timestep from flowing into future models, aligning with the characteristics of time-series data.

At the beginning of each training epoch, $M_0$ is initialized to $F_0$ for all $\alpha$, enhancing stability for subsequent EMA accumulation. Since $SA$ is proposed to address long-range dependencies in training, it lacks sufficient information in the early stages when not enough steps have been seen. Therefore, for the stability of training, we linearly warm up the learning rate for the first $1/(1-max (\textrm{\textit{smoothing factors}}))$ timesteps at the beginning of each training epoch. 
The overall training of both the base model and $BSA$ is conducted according to Algorithm~\ref{alg:1}.

\begin{wrapfigure}{h}{0.54\textwidth} 
\vspace{-25pt}
\begin{minipage}{0.53\textwidth}
\begin{algorithm}[H]
   \caption{ Batched Spectral Attention (1 epoch)}
   \label{alg:1}
\begin{algorithmic}
\footnotesize
\STATE {\bfseries Input:} Trained up to ($E$-$1$)th epoch \\
TSF model $f_{\theta}$, 
$BSA$ with \textit{sa-matrix}, \textit{smoothing factors}\\
Train data: $\mathfrak{D}_{tr} = \left \{ (X_0, Y_0),..., (X_{tr}, Y_{tr}) \right \}$\\
Valid data: $\mathfrak{D}_{val} = \left \{ (X_{tr+1}, Y_{tr+1}),..., (X_{val}, Y_{val}) \right \}$\\
mini-batch size: B

\STATE {\bfseries Train:} \\
$st = 0, ed = B-1$
\STATE initialize $M_0$ in $BSA$ with $X_0$ and $f_{\theta}$ ($:=f_{2\theta}\cdot f_{1\theta}$)
\FOR[: \textit{train phase}]{$X_{\left [st,ed\right ]}, Y_{\left [st,ed\right ]}\textrm{ in } \mathfrak{D}_{tr}$ }
\STATE $P_{\left [st,ed\right ]} \leftarrow f_{2\theta} \cdot BSA \cdot f_{1\theta}(X_{\left [st,ed\right ]})$

\STATE  $\mathcal{L} = \mathcal{L}_\text{mse}(P_{\left [st,ed\right ]},Y_{\left [st,ed\right ]})$
\STATE Compute $\nabla\mathcal{L}$ and adjust learning rate
\STATE update $f_{\theta}$, \textit{sa-matrix}, \textit{smoothing factors} 
\STATE $st += B$, $ed = min(ed+B, tr)$
\ENDFOR

$st = tr+1, ed = tr+B$
\FOR [: \textit{Validation phase}]{$X_{\left [st,ed\right ]}, Y_{\left [st,ed\right ]}\textrm{ in } \mathfrak{D}_{val}$}
\STATE $P_{\left [st,ed\right ]} \leftarrow f_{2\theta} \cdot BSA \cdot f_{1\theta}(X_{\left [st,ed\right ]})$
\STATE $\mathcal{L} = \mathcal{L}_\text{mse}(P_{\left [st,ed\right ]},Y_{\left [st,ed\right ]})$
\STATE accumulate $\mathcal{L}$ and calculate validation loss $\mathcal{L}_{val}$
\STATE $st += B$, $ed = min(ed+B, val)$
\ENDFOR
\STATE {\bfseries Output:} Trained up to $E$ th epoch\\
$f_{\theta}$, \textit{sa-matrix}, $\left \{\alpha_{1},...,\alpha_{K}\right \}$, $\mathcal{L}_{val}$ for $E$ th epoch
\end{algorithmic}
\end{algorithm}
\vspace{-45pt}
\end{minipage}
\end{wrapfigure}

In SA, the \textit{smoothing factors} $\left \{\alpha_{1},...,\alpha_{K}\right \} \\\in \mathbb{R}^{K}$ were given as scalar values, whereas in $BSA$, they are expressed by learnable parameters. This is because $BSA$ can utilize additional past information in training beyond the look-back window by incorporating a batch-sized time window, allowing it to determine the extent of long-range dependency required for training.
To keep the smoothing factors between 0 and 1, we initialized learnable parameters by applying an inverse sigmoid to the initial smoothing factors and then applied a sigmoid function in training.

So far, we assume the feature $F$ from the base model as a vector. However, the output of the intermediate layers of the model is often represented as a tensor with two or more dimensions. In real practice, we use additional channel dimensions in $BSA$ to process the activation tensor, which acts as applying multiple $BSA$ modules simultaneously.

\subsection{Consecutive dataset split}
In the TSF scenario, the entire time series data $\left \{ x_1,..., x_{T_{end}} \right \}$ is divided into train, validation, and test sets in chronological order. Let $T_{tr}$ and $T_{val}$ denote the last time in the train and validation data respectively. Model training occurs using data for t in [1, $T_{tr}$], while model selection for the best model can utilize data for t in [$T_{tr+1}$, $T_{val}$]. The test set comprises predicting time steps [$T_{val+1}$, $T_{end}$], which are not accessible during training or validation.
However, since each training sample consists of the look-back window of size L and a prediction window of size S, the training input samples are restricted to $X_{\left [L, T_{tr}-S\right ]}$. 
Validation samples and test input samples range from $X_{\left [T_{tr}, T_{val}-S\right ]}$, and from $X_{\left [T_{val}, T_{end}-S\right ]}$, respectively. While this approach is plausible for independent data like images, it is unnatural for sequential data, as it leaves unreachable gaps ($X_{\left [T_{tr}-S+1, T_{tr}-1\right ]}$, $X_{\left [T_{val}-S+1, T_{val}-1\right ]}$), undermining the consecutive characteristics.
We filled the missing gaps, making train, validation, and test sets consecutive so that our $BSA$ model could update momentum continuously. For fair evaluation, added samples were not used for either model training, validation, or performance assessment. 
Detailed explanations of model validation and evaluation are provided in Appendix~\ref{appen:A.3}.
The full code is available at \href{https://github.com/DJLee1208/BSA\_2024}{https://github.com/DJLee1208/BSA\_2024}.


\section{Experiments}
\label{experiments}
We first evaluate BSA using state-of-the-art TSF models and various real-world time series forecasting scenarios in section \ref{main_result}. 
Next, to demonstrate that BSA effectively addresses long-range dependencies and is robust to distribution shift, we perform experiments on synthetic signals of various frequencies in section \ref{synthetic_data}.
Finally, we analyze the BSA's performance variations depending on the insertion sites within the base model, examine computation and memory costs, and conduct an ablation study in section \ref{ablation}.

\textbf{Datasets. }
We use eleven real-world public datasets: Weather, Traffic, ECL, ETT (4 sub-datasets; h1, h2, m1, m2), Exchange, PEMS03, EnergyData, and Illness~\cite{EnergyData, LSTNet, SCINet, Autoformer}. In the Illness dataset, the look-back window is set to 36, and the forecasting lengths are 24, 36, 48, and 60. For the other datasets, the look-back window is set to 96, and the forecasting lengths are 96, 192, 336, and 720. Train, validation, and test split ratio are 0.6, 0.2, 0.2 for the ETT dataset and 0.7, 0.1, 0.2 for the Weather, Traffic, ECL, Exchange, PEMS03, EnergyData, and Illness datasets. The Weather, Traffic, and ECL datasets settings follow the TimesNet paper~\cite{Timesnet}. Details on datasets are provided in Appendix \ref{appen:A.1}.

\textbf{Baseline models. }
As a benchmark, we apply BSA to 7 recent or well-known forecasting models. (1) Linear based models: DLinear~\cite{DLinear}, RLinear~\cite{RLinear}, FreTS~\cite{FreTS} (2) Convolution based methods: TimesNet~\cite{Timesnet} (3) Transformer based methods: iTransformer~\cite{iTransformer}, Crossformer~\cite{Crossformer}, PatchTST~\cite{PatchTST}. For each dataset, the model structure configuration is based on the Time-Series-Library~\cite{Timesnet}. Details on the base model configurations are provided in Appendix \ref{appen:A.2}.

\textbf{Training details. }
We first train the base model for more than 30 epochs (20 epochs for the Traffic dataset) using Adam~\cite{Adam} to ensure that the validation MSE saturates, while also conducting an extensive hyperparameter search. Then, we fine-tune with the BSA module attached to the pre-trained base model. All experiments are based on the average values of three random seeds. We provide further details in Appendix~\ref{appen:A.3}.

\subsection{Real world datasets}
\label{main_result}
\begin{table}
\centering
\caption{Forecasting results averaged across prediction lengths $S \in \{96, 192, 336, 720\}$ and three seeds. Illness and Exchange datasets use different prediction lengths due to short data lengths. Higher performance between the base model and BSA is bolded. The Avg. column shows the mean across all datasets. Red indicates p-value < 0.05 in paired t-test. Abbreviations are as follows, We.; Weather, Tr.: Traffic, Eh1: ETTh1, Eh2: ETTh2, Em1: ETTm1, Em2: ETTm2, Ex.:Exchange, PE.: PEMS03, En.: EnergyData, Il.: Illness. Full results are provided in Appendix \ref {appen:result1-1}.}
\vspace{5pt}
\label{tab:short_main}
\resizebox{1.0\textwidth}{!}{%
\begin{tabular}{ccc|ccccccccccc|c}
\toprule
\multicolumn{2}{c}{Model}& Metric               & We.            & Tr.            & ECL            & Eh1            & Eh2            & Em1            & Em2            & Ex.            & PE.             & En.            & Il.            & Avg.           \\
\midrule
\multirow{4.5}{*}{DLinear}      & \multirow{2}{*}{base} & MSE & 0.244          & 0.737          & 0.227          & 0.529          & 0.349          & 0.462          & 0.248          & 0.088          & 0.436          & 0.874          & 2.849          & 0.640          \\
                              &                       & MAE & 0.308          & 0.451          & 0.325          & 0.507          & 0.406          & 0.447          & 0.332          & 0.211          & 0.502          & 0.683          & \textbf{1.111} & 0.480          \\
                              \cmidrule{2-15} 
                              & \multirow{2}{*}{BSA}  & MSE & \textbf{0.220} & \textbf{0.691} & \textbf{0.217} & \textbf{0.527} & \textbf{0.341} & \textbf{0.441} & \textbf{0.221} & \textbf{0.081} & \textbf{0.384} & \textbf{0.838} & \textbf{2.797} & {\color[HTML]{ff0000}\textbf{0.614}} \\
                              &                       & MAE & \textbf{0.281} & \textbf{0.435} & \textbf{0.316} & \textbf{0.507} & \textbf{0.398} & \textbf{0.447} & \textbf{0.310} & \textbf{0.207} & \textbf{0.464} & \textbf{0.667} & 1.114          & {\color[HTML]{ff0000}\textbf{0.468}} \\
                              \midrule
\multirow{4.5}{*}{RLinear}      & \multirow{2}{*}{base} & MSE & 0.248          & 0.723          & 0.231          & 0.551          & 0.309          & 0.488          & 0.220 & 0.085          & 0.992          & 0.816          & 2.637          & 0.664          \\
                              &                       & MAE & 0.274          & 0.442          & 0.319          & \textbf{0.516} & 0.370          & 0.457          & 0.307          & 0.204          & 0.740          & 0.630          & 1.013          & 0.479          \\
                              \cmidrule{2-15} 
                              & \multirow{2}{*}{BSA}  & MSE & \textbf{0.237} & \textbf{0.665} & \textbf{0.212} & \textbf{0.545} & \textbf{0.304} & \textbf{0.475} & \textbf{0.210} & \textbf{0.085} & \textbf{0.674} & \textbf{0.797} & \textbf{2.605} & {\color[HTML]{ff0000}\textbf{0.619}} \\
                              &                       & MAE & \textbf{0.267} & \textbf{0.404} & \textbf{0.301} & 0.517          & \textbf{0.370} & \textbf{0.453} & \textbf{0.303} & \textbf{0.203} & \textbf{0.607} & \textbf{0.622} & \textbf{1.005} & {\color[HTML]{ff0000}\textbf{0.459}} \\
                              \midrule
\multirow{4.5}{*}{FreTS}        & \multirow{2}{*}{base} & MSE & 0.240          & 0.527          & 0.195          & 0.513          & 0.310          & 0.460          & \textbf{0.234} & 0.095          & 0.259          & 0.976          & 5.073          & 0.807          \\
                              &                       & MAE & \textbf{0.286} & 0.322          & \textbf{0.285} & 0.506          & 0.393          & 0.450          & \textbf{0.321} & 0.232          & 0.354          & 0.735          & 1.603          & 0.499          \\
                              \cmidrule{2-15} 
                              & \multirow{2}{*}{BSA}  & MSE & \textbf{0.236} & \textbf{0.502} & \textbf{0.192} & \textbf{0.511} & \textbf{0.308} & \textbf{0.431} & 0.241          & \textbf{0.089} & \textbf{0.229} & \textbf{0.940} & \textbf{4.930} & {\color[HTML]{ff0000}\textbf{0.783}} \\
                              &                       & MAE & 0.292          & \textbf{0.317} & 0.287          & \textbf{0.505} & \textbf{0.387} & \textbf{0.442} & 0.326          & \textbf{0.219} & \textbf{0.324} & \textbf{0.710} & \textbf{1.591} & {\color[HTML]{ff0000}\textbf{0.491}} \\
                              \midrule
\multirow{4.5}{*}{TimesNet}     & \multirow{2}{*}{base} & MSE & 0.260          & \textbf{0.620} & 0.203          & 0.576          & 0.386          & 0.508          & 0.247          & 0.094          & \textbf{0.232} & 0.880          & \textbf{2.719} & 0.611          \\
                              &                       & MAE & 0.284          & \textbf{0.333} & 0.301          & 0.545          & 0.427          & 0.482          & 0.326          & 0.219          & \textbf{0.311} & 0.662          & 0.942          & 0.439          \\
                              \cmidrule{2-15} 
                              & \multirow{2}{*}{BSA}  & MSE & \textbf{0.252} & 0.624          & \textbf{0.199} & \textbf{0.569} & \textbf{0.378} & \textbf{0.499} & \textbf{0.229} & \textbf{0.094} & 0.234          & \textbf{0.861} & 2.720          & {\color[HTML]{ff0000}\textbf{0.605}} \\
                              &                       & MAE & \textbf{0.279} & 0.335          & \textbf{0.298} & \textbf{0.540} & \textbf{0.418} & \textbf{0.476} & \textbf{0.314} & \textbf{0.219} & 0.314          & \textbf{0.655} & \textbf{0.938} & {\color[HTML]{ff0000}\textbf{0.435}} \\
                              \midrule
\multirow{4.5}{*}{iTransformer} & \multirow{2}{*}{base} & MSE & 0.256          & \textbf{0.428} & 0.181          & 0.542          & 0.321 & 0.466          & 0.224          & 0.091          & 0.262          & 0.833          & \textbf{2.454} & 0.551          \\
                              &                       & MAE & 0.277          & \textbf{0.284} & 0.272          & 0.518          & 0.380          & 0.454          & 0.314          & 0.211          & 0.345          & 0.639          & \textbf{0.947} & 0.422          \\
                              \cmidrule{2-15} 
                              & \multirow{2}{*}{BSA}  & MSE & \textbf{0.236} & 0.428          & \textbf{0.176} & \textbf{0.537} & \textbf{0.317} & \textbf{0.466} & \textbf{0.219} & \textbf{0.090} & \textbf{0.198} & \textbf{0.786} & 2.540          & \textbf{0.545} \\
                              &                       & MAE & \textbf{0.268} & 0.288          & \textbf{0.270} & \textbf{0.517} & \textbf{0.378} & \textbf{0.452} & \textbf{0.310} & \textbf{0.211} & \textbf{0.298} & \textbf{0.619} & 0.957          & {\color[HTML]{ff0000}\textbf{0.415}} \\
                              \midrule
\multirow{4.5}{*}{Crossformer}  & \multirow{2}{*}{base} & MSE & 0.240          & 0.565          & \textbf{0.182} & \textbf{0.517} & 0.323          & 0.485          & \textbf{0.243} & 0.220          & 0.212          & \textbf{1.178} & 4.946          & 0.828          \\
                              &                       & MAE & 0.292          & 0.292          & \textbf{0.276} & \textbf{0.520} & 0.402          & 0.470          & 0.342          & 0.345          & 0.299          & \textbf{0.817} & 1.518          & 0.507          \\
                              \cmidrule{2-15} 
                              & \multirow{2}{*}{BSA}  & MSE & \textbf{0.227} & \textbf{0.554} & 0.186          & 0.520          & \textbf{0.317} & \textbf{0.468} & 0.245          & \textbf{0.219} & \textbf{0.201} & 1.204          & \textbf{4.809} & \textbf{0.814} \\
                              &                       & MAE & \textbf{0.284} & \textbf{0.288} & 0.280          & 0.522          & \textbf{0.396} & \textbf{0.469} & \textbf{0.340} & \textbf{0.340} & \textbf{0.290} & 0.823          & \textbf{1.489} & {\color[HTML]{ff0000}\textbf{0.502}} \\
                              \midrule
\multirow{4.5}{*}{PatchTST}     & \multirow{2}{*}{base} & MSE & 0.255          & \textbf{0.467} & 0.198          & \textbf{0.535} & 0.317          & 0.473          & 0.220          & 0.087          & 0.361          & 0.839          & 2.016          & 0.524          \\
                              &                       & MAE & 0.278          & \textbf{0.296} & 0.286          & \textbf{0.516} & 0.375          & 0.456          & 0.310          & 0.204          & 0.413          & 0.642          & 0.880          & 0.423          \\
                              \cmidrule{2-15} 
                              & \multirow{2}{*}{BSA}  & MSE & \textbf{0.236} & 0.467          & \textbf{0.189} & 0.539          & \textbf{0.314} & \textbf{0.458} & \textbf{0.214} & \textbf{0.086} & \textbf{0.244} & \textbf{0.788} & \textbf{1.974} & {\color[HTML]{ff0000}\textbf{0.501}} \\
                              &                       & MAE & \textbf{0.268} & 0.297          & \textbf{0.285} & 0.520          & \textbf{0.374} & \textbf{0.451} & \textbf{0.306} & \textbf{0.202} & \textbf{0.341} & \textbf{0.621} & \textbf{0.866} & {\color[HTML]{ff0000}\textbf{0.412}}\\
\bottomrule
\end{tabular}
}
\vspace{-10pt}
\end{table}

\begin{figure}[t!]
\vspace{-15pt}
    \centering
    \begin{subfigure}[b]{0.245\columnwidth}
        \includegraphics[width=\linewidth]{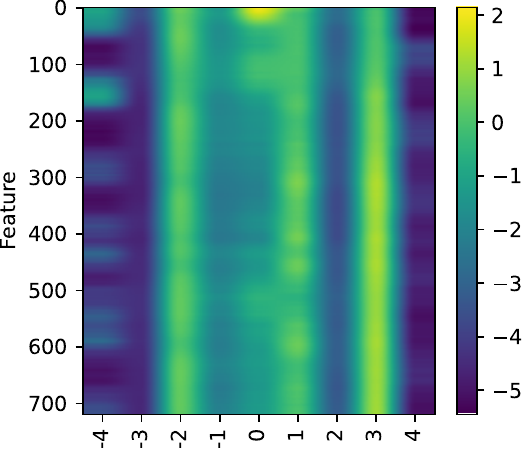}
        \caption{SA-Matrix heat map}
        \label{fig:4.1.a}
    \end{subfigure}
    \begin{subfigure}[b]{0.245\columnwidth}
        \includegraphics[width=\linewidth]{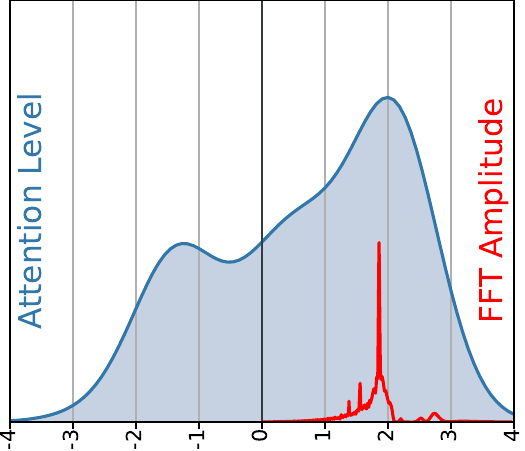}
        \caption{Weather-Temp.(°C)}
        \label{fig:4.1.b}
    \end{subfigure}
    \begin{subfigure}[b]{0.245\columnwidth}
        \includegraphics[width=\linewidth]{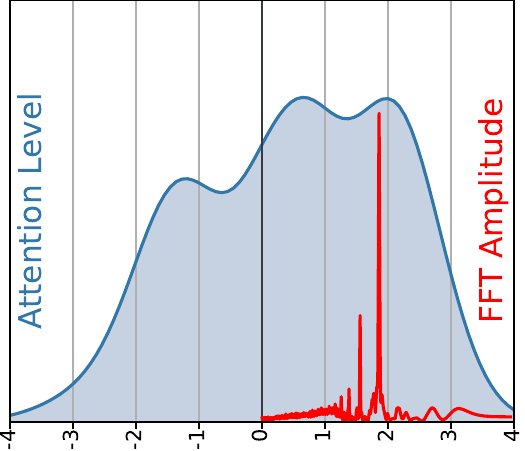}
        \caption{Weather-SWDR(W/m)}
        \label{fig:4.1.c}
    \end{subfigure}
    \begin{subfigure}[b]{0.245\columnwidth}
        \includegraphics[width=\linewidth]{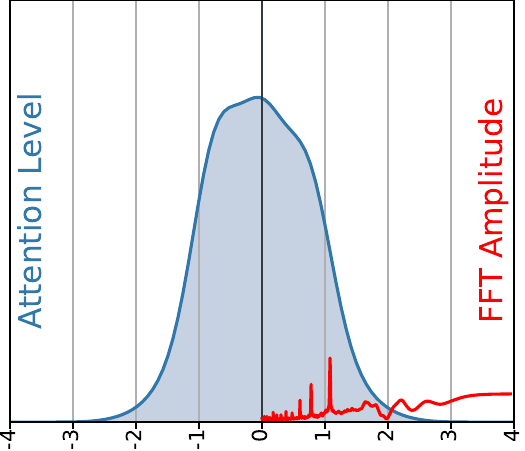}
        \caption{ETTh1-HULL}
        \label{fig:4.1.d}
    \end{subfigure}
    \vspace{-15pt}
    \caption{This figure illustrates the analysis of the SA-matrix of the DLinear model trained on the 720-step prediction task for the Weather and ETTh1 datasets. Panel (a) shows the heatmap of the SA-matrix, and (b)-(d) show the attention and FFT graphs.}
    \label{fig:4.1}
    \vspace{-10pt}
\end{figure}
We demonstrate that BSA improves forecasting performance by providing the ability to address long-range dependencies, regardless of the base model architecture. Table~\ref{tab:short_main} presents the averaged values over prediction lengths 
for both the base and the BSA applied model 
across 11 time series datasets and 7 TSF models. 
We apply the BSA module at the beginning of the base model's activation, with the number of BSA channels matching the number of channels in the data (position 1 in Figure \ref{fig:4.3.1}). BSA effectively improves MSE and MAE across all architectures.
The average performance gain, in terms of MSE, ranged from as low as 0.96\% to as high as 7.2\%, with linear-based models demonstrating relatively high performance.
Paired t-tests demonstrate statistically significant (p-value < 0.05) improvements in most MSE and all MAE across various models.
This result emphasizes the model-agnostic versatility of BSA.
Furthermore, BSA exhibited overall performance gain across all datasets, with the enhancement being statistically significant in 82\% of cases.
BSA's higher gain on ETTm compared to ETTh, both derived from the same signal but with different sampling rates, further indicates BSA's effectiveness in handling long-range dependencies.

To understand how BSA addresses long-range dependencies, we conduct an internal inspection. In Figure~\ref{fig:4.1.a}, we present the heat map of the trained SA-Matrix of the DLinear (Temperature(°C) channel, Weather data). The positive x-axis corresponds to the $log_{10}$ values of the periods preserved by the low-pass filter, derived with the smoothing factor. Negative values correspond to high-frequency components. The blue graph in Figure~\ref{fig:4.1.b} represents 
the SA-Matrix averaged over the feature dimension, 
illustrating the frequencies to which BSA attends overall. 
The red graph represents the result of applying the Fast Fourier Transform (FFT) and denoising to the raw signal.
The blue graph skewed towards the low-frequency side indicates that the BSA effectively captures the long-range trend of the data. Figure~\ref{fig:4.1.c} depicts the graph for the SWDR (Short Wave Downward Radiation per unit area, W/m) channel in the same SA-Matrix. While not identical to Figure~\ref{fig:4.1.b}, it also exhibits strong attention towards the low-frequency pattern. In contrast, Figure~\ref{fig:4.1.d}, the FFT graph for the HULL (High UseLess Load channel, ETTh1 data), shows that the signal itself lacks long-range trends, resulting in the symmetric SA-Matrix. This result demonstrates that BSA operates as intended, learning the low-frequency components of the signal for future prediction. 
We provided other graphs and detailed information on the graph plotting method in Appendix~\ref{appen:result1-2}.

\subsection{Synthetic datasets}
\label{synthetic_data}
\begin{figure}[b!]
\vspace{-5pt}
    \centering
    \includegraphics[width=0.80\linewidth]{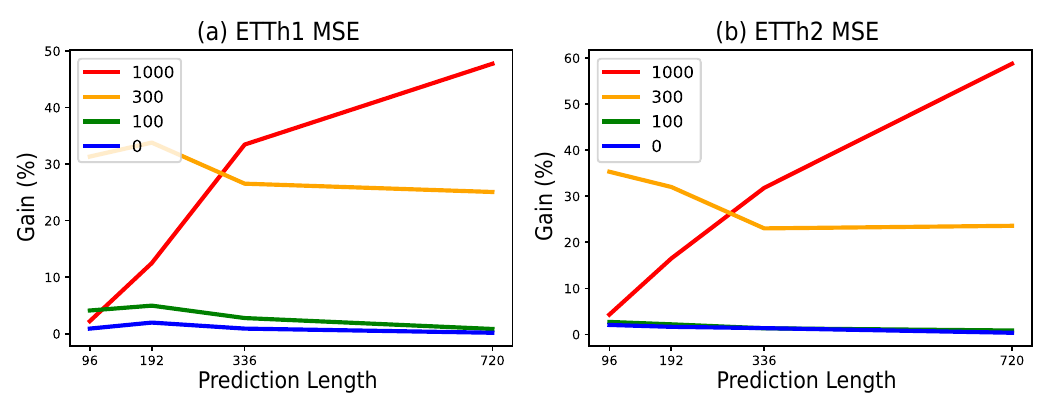}
    \vspace{-2pt}
    \caption{Results of the iTransformer model on synthetic (a) ETTh1 and (b) ETTh2 datasets. The x-axis is the prediction length (96, 192, 336, 720), and the y-axis is the performance improvement (\%) compared to the base model. Each color represents the different periods of the sine wave added to the natural data. 0 indicates original data and serves as the baseline.}
    \label{fig:4.2.1}
    \vspace{-10pt}
\end{figure}

To further demonstrate that BSA learns long-range dependencies beyond the look-back window, 
we add sine waves with periods of 100, 300, and 1000 to the natural data while maintaining the mean and standard deviation (Refer to Appendix~\ref{appen:result2} for details on synthetic data generation). 
Figure~\ref{fig:4.2.1} illustrates the performance improvement of BSA over the iTransformer model on the ETTh1 and ETTh2 datasets. The x-axis is the prediction length, and each line represents the period of the added sine wave. Performance improvement is calculated as 100$\times$(base MSE - BSA MSE) / base MSE. 
While the base model with a 96-length look-back window is expected to learn the 100-period trend, BSA outperformed it, especially for 96 and 192-step predictions (green line).
The yellow line (period 300) shows nearly a 30\% performance improvement across all prediction lengths. While the base model fails to learn the long-range interactions within a period of 300, BSA captures and utilizes the underlying trend.
BSA also learns the 1000-period signal (red line) and demonstrates substantial improvements, especially in long prediction-length (336, 720) tasks.
These results show that BSA effectively learns long-range patterns beyond the look-back window, essential for future prediction.

\begin{figure}[t!]
\vspace{-10pt}
    \centering
    \begin{subfigure}[b]{0.245\columnwidth}
        \includegraphics[width=\linewidth]{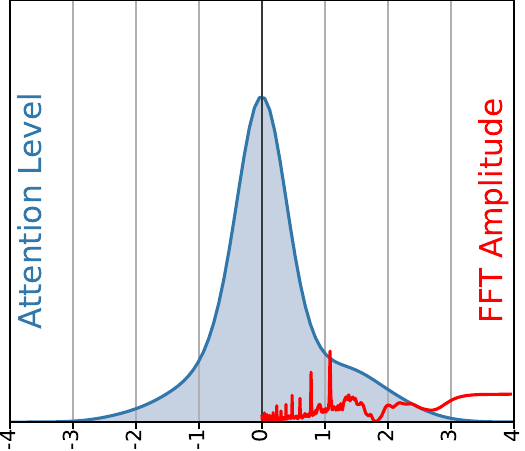}
        \caption{LUFL original}
        \label{fig:4.2.2-a}
    \end{subfigure}
    \begin{subfigure}[b]{0.245\columnwidth}
        \includegraphics[width=\linewidth]{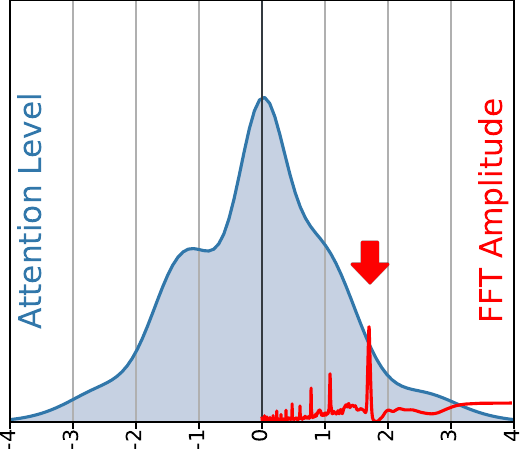}
        \caption{LUFL period 100}
        \label{fig:4.2.2-b}
    \end{subfigure}
    \begin{subfigure}[b]{0.245\columnwidth}
        \includegraphics[width=\linewidth]{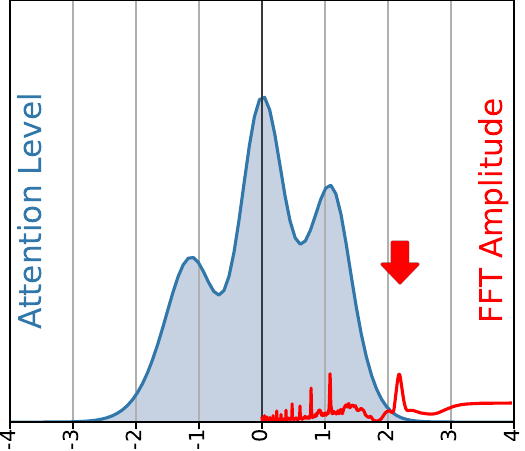}
        \caption{LUFL period 300}
        \label{fig:4.2.2-c}
    \end{subfigure}
    \begin{subfigure}[b]{0.245\columnwidth}
        \includegraphics[width=\linewidth]{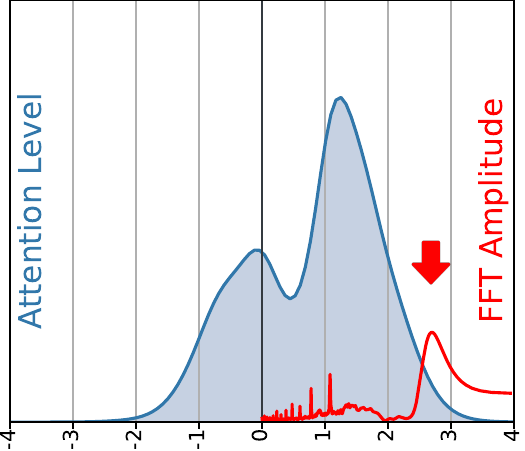}
        \caption{LUFL period 1000}
        \label{fig:4.2.2-d}
    \end{subfigure}
    \vspace{-10pt}
    \caption{Attention and FFT graphs on LUFL channel of the original and synthetic ETTh1 data (iTransformer, 720-step prediction). (a) is from the original data and (b)-(d) are from the synthetic data created by adding sine waves with periods of 100, 300, and 1000, respectively. The red arrows on the FFT graphs show the added synthetic signals. Full visualization is provided in Appendix~\ref{appen:result3}}
    \label{fig:4.2.2}
    \vspace{-5pt}
\end{figure}

Figure~\ref{fig:4.2.2} is generated from LUFL (Low UseFul Load) channel of ETTh1 data and with added sine waves of periods 100, 300, and 1000. The red arrow shows the added synthetic trend in the FFT graph (red line).
As long-range trends are introduced, the attention graph (blue line) shifts from resembling symmetric Gaussian to a low-frequency bias. Longer sine wave periods cause greater shifts, prioritizing long-range information for predictions.

\subsection{Analysis and ablation studies}
\label{ablation}

\vspace{-5pt}
\begin{minipage}[t!]{\textwidth}
\begin{minipage}[b]{0.58\textwidth}
\centering
\resizebox{0.8\textwidth}{!}{%
\begin{tabular}{@{}r|cccc@{}}
\toprule
\multicolumn{1}{c|}{\multirow{2}{*}{Location}} & \multicolumn{2}{c}{iTransformer} & \multicolumn{2}{c}{DLinear} \\
\multicolumn{1}{c|}{} & MSE & MAE & MSE & MAE \\ \midrule
\rowcolor{LightCyan} 
1 & 0.2357 & 0.2682 & \textbf{0.2196} & \textbf{0.2815} \\
2 & 0.2352 & 0.2681 & - & - \\
3 & 0.2329 & \textbf{0.2654} & - & - \\
4 & 0.2508 & 0.2752 & 0.2327 & 0.2994 \\
Query-5 & \textbf{0.2326} & 0.2671 & - & - \\
Key-6 & 0.2538 & 0.2762 & - & - \\
Value-7 & 0.2415 & 0.2702 & - & - \\ \midrule
baseline & 0.2556 & 0.2766 & 0.2444 & 0.3084 \\ \bottomrule
\end{tabular}%
}
\captionof{table}{Performance analysis on BSA insertion site. Each number corresponds with the insertion site in Figure~\ref{fig:4.3.1}.}
\label{tab:4.3.1}
\end{minipage}
\hfill
\begin{minipage}[b]{0.38\textwidth}
\centering
    \includegraphics[width=\linewidth]{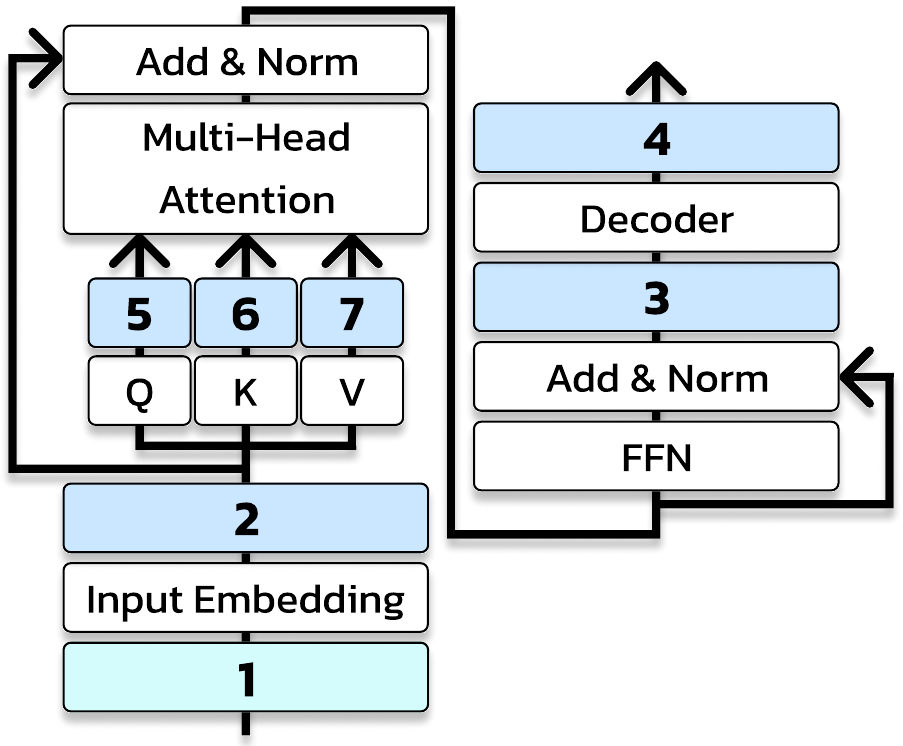}
\captionof{figure}{Schematic diagram of the BSA insertion site on Transformer.}
\label{fig:4.3.1}
\end{minipage}
\end{minipage}
\vspace{5pt}

The BSA module offers high flexibility as it can be applied to arbitrary activations of the base model. In Table~\ref{tab:4.3.1}, we analyzed the performance changes by applying BSA at various locations within the model. Each location corresponds to a number in the Transformer architecture depicted in Figure~\ref{fig:4.3.1}. While we uniformly applied BSA to position 1 for the main result Table~\ref{tab:short_main}, the results in Table~\ref{tab:4.3.1} suggest the potential for further performance enhancement by applying BSA at appropriate locations.
Additionally, while there is variability depending on the placement, the performance consistently remains higher compared to the baseline, demonstrating the stability of our method.
We provided a full result in Appendix~\ref{appen:result4.1}.

BSA shows consistent performance improvement across varying look-back window (input) lengths. Table~\ref{tab:input_length} demonstrates BSA's superiority for look-back window lengths of 48, 96, and 192. Notably, while the baseline model's performance drops significantly with shorter inputs, BSA maintains high performance. 

\begin{minipage}[h!]{\textwidth}
\begin{minipage}[h]{0.51\textwidth}
\centering
\captionof{table}{BSA's performance improvement (\%, denoted with *) compared to base model across three input lengths $\{48, 96, 192\}$ (iTransformer, average over 4 prediction lengths and 3 seeds). The full result, including other models, is in Appendix~\ref{appen:result4.2}.}
\vspace{1pt}
\resizebox{1.\textwidth}{!}{
\begin{tabular}{c|cccc}
& \multicolumn{2}{c}{Weather}& \multicolumn{2}{c}{PEMS03}\\
\multirow{-2}{*}{\textit{\begin{tabular}[c]{@{}c@{}}Input-\\ length\end{tabular}}} & \cellcolor[HTML]{EFEFEF}MSE*(\%) & \cellcolor[HTML]{EFEFEF}MAE*(\%) & \cellcolor[HTML]{EFEFEF}MSE*(\%) & \cellcolor[HTML]{EFEFEF}MAE*(\%)\\ \hline
48& 12.08 & 5.21& 29.45& 17.35\\
\rowcolor[HTML]{EFEFEF} 96& 7.78 & 3.03& 24.55& 13.66\\
192& 6.87 & 3.34& 9.63& 5.18                  
\end{tabular}
}
\label{tab:input_length}
\end{minipage}
\hfill
\vspace{-3pt}
\begin{minipage}[h]{0.45\textwidth}
\vspace{-3pt}
\centering
\captionof{table}{Computational cost increase with the BSA (\%), averaged over 3 seeds and 4 prediction lengths on the PEMS03 dataset. TN: TimesNet, iTF: iTransformer, CF: Crossformer, PTST: PatchTST. The full result, including other datasets, is in Appendix~\ref{appen:result4.3}.}
\vspace{-2pt}
\resizebox{0.9\textwidth}{!}{
\begin{tabular}{l|cccc}
          Increase (\%) & TN & iTF & CF & PTST \\ \hline
\rowcolor[HTML]{EFEFEF} 
{\color[HTML]{333333} Time} & {\color[HTML]{333333} 0.21} & {\color[HTML]{333333} 2.24} & {\color[HTML]{333333} -2.04} & {\color[HTML]{333333} -0.29} \\
Memory    & 0.34     & 2.34         & 0.17        & 0.38     \\
\rowcolor[HTML]{EFEFEF} 
Parameter & 0.16     & 4.88         & 1.96        & 2.25    
\end{tabular}
\label{tab:computation}
}
\end{minipage}
\vspace{10pt}
\end{minipage}

We also demonstrate the impact of using BSA on the training time, peak memory, and the number of model parameters. The extra computation required for BSA is constant with data length and linear with a look-back window length (c.f. quadratic for the base model with transformer architecture). Table~\ref{tab:computation} demonstrates the computational cost of the BSA is quite small, showing less than a 5\% increase even with the large PEMS03 dataset. 

\begin{wraptable}{r}{6.3cm}
\vspace{-12pt}
\caption{Ablation study with 3 components in BSA (Weather data, 720-step prediction, iTransformer).}
\vspace{-5pt}
\centering
\resizebox{0.45\textwidth}{!}{
\begin{tabular}{ccc|ll}
BPTT & SFs & Learn SF & \multicolumn{1}{c}{MSE} & \multicolumn{1}{c}{MAE} \\ 
\hline
\rowcolor{LightGray}\multicolumn{3}{c|}{baseline}  & \multicolumn{1}{c}{0.3551}    & \multicolumn{1}{c}{0.3473}    \\
\hline
     &          &            & \multicolumn{1}{c}{0.3480}    & \multicolumn{1}{c}{0.3452}    \\
\rowcolor{LightGray}\checkmark    &          &            & \multicolumn{1}{c}{0.3401}    & \multicolumn{1}{c}{0.3452}    \\
    \checkmark & \checkmark  &            & \multicolumn{1}{c}{0.3320}    & \multicolumn{1}{c}{0.3395}    \\
\rowcolor{LightGray} \checkmark &  \checkmark & \checkmark &  \textbf{0.3263}	 &   \textbf{0.3358}
\end{tabular}
\label{tab:albation}
}
\vspace{-20pt}
\end{wraptable}

We conduct an ablation study on the three key components that constitute BSA (Table~\ref{tab:albation}). ``BPTT'' refers to whether gradients can flow between samples within the mini-batch. Without BPTT, BSA learns similarly to SA. ``SFs'' denotes whether to use multiple smoothing factors.
Lastly, ``Learn SF'' indicates whether the smoothing factor is treated as learnable. 
The results indicate that each component significantly contributes to the performance improvement of BSA.

\section{Conclusion}
\label{conclusion}
Our study addresses the challenges in handling long-range dependencies inherent in time series data by introducing a fast and effective Spectral Attention mechanism. By preserving temporal correlations and enabling the flow of gradients between samples, this mechanism facilitates the model in capturing crucial long-range interactions essential for accurate future predictions. Therefore, our research paves the way for fixed-sized input models to effectively handle long-range dependencies extending far beyond the input window.
Through extensive experimentation, we demonstrated that our Spectral Attention mechanism enhances performance across various base architectures, with its ability to grasp long-term dependencies being the key factor behind this improvement.
BSA effectively tackles long-term fluctuations, complementing the base model's capacity to manage intricate yet short-term patterns. This integrated model holds promise for improving real-world application performance. For instance, it could boost weather forecast accuracy by simultaneously capturing minute-by-minute weather changes and seasonal variations. Moreover, when predicting deterioration from a patient's real-time data, it can consider medications with lengthy onset times.
Our study has limitations: we did not analyze the impact of BSA placement within the base model in detail. Also, BSA's performance gains may be limited when applied to datasets with only high-frequency information within the look-back window. 
These issues should be addressed in future research.

\section{Acknowledgement}
This research was supported by a grant of the MD-Phd/Medical Scientist Training Program through the Korea Health Industry Development Institute (KHIDI), funded by the Ministry of Health \& Welfare, Republic of Korea, National Research Foundation of Korea (NRF) grants funded by the Korea government (Ministry of Science and ICT, MSIT) (2022R1A3B1077720 and 2022R1A5A708390811), Institute of Information \& Communications Technology Planning \& Evaluation (IITP) grants funded by the Korea government (MSIT) (2021-0-01343: Artificial Intelligence Graduate School Program (Seoul National University), 2022-0-00959 and IITP-2024-RS-2024-00397085: Leading Generative AI Human Resources Development), and the BK21 FOUR program of the Education and Research Program for Future ICT Pioneers, Seoul National University in 2024, AI-Bio Research Grant through Seoul National University, Hyundai Motor Company, HUINNO AIM
Company through HA-Rnd-2325-PredictClinicalDeterioration.

\clearpage
\bibliography{references}
\bibliographystyle{plain}

\clearpage
\appendix

\section{Details on Datasets, Models, and Training}
\label{appen:A}
\subsection{Details on datasets} 
\label{appen:A.1}
\textbf{Dataset Information.} 
We conducted experiments on 11 real-world datasets to assess the performance of baseline models and the proposed BSA method. 
The Weather dataset~\cite{Autoformer} includes 21 meteorological factors acquired every 10 minutes in 2020 from the Weather Station of the Max Planck Institute for Biogeochemistry. 
The Traffic dataset~\cite{Autoformer}  records hourly road occupancy rates from 862 sensors on San Francisco Bay Area freeways, covering the period from January 2015 to December 2016. 
The ECL dataset~\cite{Autoformer} captures the hourly electricity consumption of 321 clients.
The ETT dataset~\cite{Informer} contains seven factors related to electricity transformers, spanning from July 2016 to July 2018. It is divided into four sub-datasets: ETTh1 and ETTh2 are recorded hourly, while ETTm1 and ETTm2 are collected every 15 minutes. 
The Exchange dataset~\cite{Autoformer} comprises panel data of daily exchange rates from eight countries, ranging from 1990 to 2016. 
Illness dataset~\cite{Autoformer} contains weekly data on influenza-like illness (ILI) cases recorded by the U.S. Centers for Disease Control and Prevention (CDC) from 2002 to 2021. The dataset tracks the proportion of ILI patients relative to the overall number of patients seen during that period.
PEMS03 dataset~\cite{SCINet} is a sub-dataset of the PEMS dataset, which includes public traffic network data from California, recorded at 5-minute intervals.
The EnergyData dataset~\cite{EnergyData} comprises hourly end-use measurements gathered from 454 residential properties and 140 commercial establishments located in the Pacific Northwest.
All these public datasets were downloaded from the referenced sources in March 2024.

\textbf{Dataset split.} 
We adhere to the data processing protocol and train-validation-test split used in TimesNet~\cite{Timesnet}, where the training, validation, and test datasets are sequentially separated in chronological order. Our paper's data split ratios for train, validation, and test set are as follows: (0.7, 0.1, 0.2) for Weather, Traffic, ECL, Exchange, Ilnness, PEMS03, and EnergyData datasets and (0.6, 0.2, 0.2) for ETT. The details of datasets are provided in Table ~\ref{tab:dataset}. 

\textbf{Forecasting setting.} 
Following the approach in iTransformer~\cite{iTransformer}, the look-back window length is set to \{96\}, while the forecast lengths are \{96, 192, 336, 720\} for the Weather, Traffic, ECL, ETT, PEMS03, and EnergyData datasets. Forecasting lengths is only \{96\} for the Exchange dataset since the dataset is too short, which causes biased best model selection with the validation set. For Illness dataset, the look-back window length is \{36\} and the forecasting lengths are \{24, 36, 48, 60\}.

\begin{table}[htbp]
\centering
\caption{Summary of Datasets. \textit{Channel} denotes the number of time series variables (channels) for each dataset. \textit{Pre-Train} is the number of training epochs for baseline model saturation. \textit{Finetune} is the number of fine-tuning epochs for our method. \textit{Data Split} means the number of time steps in (Train, Validation, Test) data split respectively. \textit{Sampling Rate} denotes how often the data samples are collected. \textit{Type} is to show the domain in which the data is acquired. We have indicated cases where the data split count matches exactly with that used in TimesNet~\cite{Timesnet} by marking them with an asterisk (*). For the ETT and the Exchange dataset, the length of the downloaded data differed from the data length reported in TimesNet.}
\label{tab:dataset}
\vspace{5pt}
\resizebox{1\columnwidth}{!}{
\begin{tabular}{@{}l|cccccc@{}}
\toprule
\textbf{Dataset} & \textbf{Channel} & \textbf{Pre-Train} & \textbf{Finetune} & \textbf{Data Split} & \textbf{Sampling Rate} & \textbf{Type} \\ \midrule
Weather* & 21 & 40 & 20 & (36792, 5271, 10540) & 10min & Weather \\
Traffic* & 862 & 20 & 20 & (12185, 1757, 3509) & Hourly & Transportation \\
ECL* & 321 & 30 & 20 & (18317, 2633, 5261) & Hourly & Electricity \\
ETTh1, ETTh2 & 7 & 30 & 30 & (10357, 3485, 3485) & Hourly & Electricity \\ 
ETTm1, ETTm2 & 7 & 30 & 20 & (41713, 13937, 13937) & 15min & Electricity \\
Exchange & 8 & 30 & 30 & (5216, 761, 1518) & Daily & Economy \\
Illness & 7 & 30 & 20 & (692, 120, 226) & Weekly & Health \\
PEMS03 & 358 & 30 & 20 & (18250, 2623, 5242) & 5min & Transportation \\
EnergyData & 28 & 30 & 20 & (13719, 1975, 3948) & Hourly & Energy \\
\bottomrule
\end{tabular}
}
\end{table} 

\subsection{Details on Model Implementations}
\label{appen:A.2}
\textbf{Model configurations.}
In this section, we discuss the configurations of the baseline models and specify where the BSA module was integrated into each baseline model. For each dataset, the base model structure configuration was directly \textbf{replicated from the Time-Series-Library}~\cite{Timesnet} scripts when available. Where configurations were not provided, we adjusted them to align closely with the available examples. 

1. \textbf{DLinear}~\cite{DLinear}: The only hyperparameter for this baseline model is the (moving\_average = 25) for the series decomposition module from Autoformer~\cite{Autoformer}. We set the Individual to True so that there are separate linear models for each number of input variables.
The BSA module is implemented at the very beginning of the forward pass right before series decomposition. The BSA module is implemented for each channel, which is the number of input variables for this case.

2. \textbf{iTransformer}~\cite{iTransformer}: The hyperparameters for this baseline model are as follows: \\
(d\_model = 512, d\_ff = 512, dropout = 0.1, num\_heads = 8, encoder\_layers = 3, activation = GELU~\cite{GELU}) for the Weather, ECL, PEMS03, and EnergyData datasets, \\
(d\_model = 512, d\_ff = 512, dropout = 0.1, num\_heads = 8, encoder\_layers = 4, activation = GELU) for the Traffic dataset, \\
(d\_model = 128, d\_ff = 128, dropout = 0.1, num\_heads = 8, encoder\_layers = 2, activation = GELU) for the ETT and Exchange datasets. \\
The BSA module is implemented at the beginning part of the forward pass right after normalization from the Non-stationary Transformer~\cite{Stationary} and right before the input embedding. The BSA module is implemented for each channel, which is the number of input variables for this case.

3. \textbf{Crossformer}~\cite{Crossformer}:  The hyperparameters for this baseline model are as follows: \\
(seg\_len = 12, win\_size = 2, factor = 3, d\_model = 32, d\_ff = 32, dropout = 0.1, num\_heads = 8, encoder\_layers = 2, activation = RELU~\cite{RELU}) for the Weather and EnergyData dataset, \\
(seg\_len = 12, win\_size = 2, factor = 3, d\_model = 128, d\_ff = 128, dropout = 0.1, num\_heads = 2, encoder\_layers = 2, activation = RELU) for the Traffic dataset, \\
(seg\_len = 12, win\_size = 2, factor = 3, d\_model = 256, d\_ff = 512, dropout = 0.1, num\_heads = 8, encoder\_layers = 2, activation = RELU) for the ECL and PEMS03 dataset, \\
(seg\_len = 12, win\_size = 2, factor = 3, d\_model = 512, d\_ff = 2048, dropout = 0.1, num\_heads = 8, encoder\_layers = 2, activation = RELU) for the ETTh1 and ETTh2 datsets, \\
(seg\_len = 12, win\_size = 2, factor = 1, d\_model = 512, d\_ff = 2048, dropout = 0.1, num\_heads = 8, encoder\_layers = 2, activation = RELU) for the ETTm1 and ETTm2 datasets, \\
(seg\_len = 12, win\_size = 2, factor = 3, d\_model = 64, d\_ff = 64, dropout = 0.1, num\_heads = 8, encoder\_layers = 2, activation = RELU) for the Exchange dataset. \\
The BSA module is implemented at the very beginning of the forward pass, right before the input embedding. The BSA module is implemented for each channel, which is the number of input variables for this case.

4. \textbf{PatchTST}~\cite{PatchTST}: The hyperparameters for this baseline model are as follows: \\
(patch\_len = 16, stride = 8, d\_model = 512, d\_ff = 2048, dropout = 0.1, num\_heads = 4, encoder\_layers = 2, activation = GELU~\cite{GELU}) for the Weather and EnergyData dataset, \\
(patch\_len = 16, stride = 8, d\_model = 512, d\_ff = 512, dropout = 0.1, num\_heads = 8, encoder\_layers = 2, activation = GELU) for the Traffic dataset, \\
(patch\_len = 16, stride = 8, d\_model = 512, d\_ff = 2048, dropout = 0.1, num\_heads = 8, encoder\_layers = 2, activation = GELU) for the ECL and PEMS03 dataset,\\
(patch\_len = 16, stride = 8, d\_model = 512, d\_ff = 2048, dropout = 0.1, num\_heads = 8, encoder\_layers = 1, activation = GELU) for the ETTh1 datset,\\
(patch\_len = 16, stride = 8, d\_model = 512, d\_ff = 2048, dropout = 0.1, num\_heads = 4, encoder\_layers = 3, activation = GELU) for the ETTh2, ETTm1, and ETTm2 datsets.\\
The BSA module is implemented at the beginning part of the forward pass right after normalization from the Non-stationary Transformer~\cite{Stationary} and right before the patch embedding. The BSA module is implemented for each channel, which is the number of input variables for this case.

5. \textbf{TimesNet}~\cite{Timesnet}: The hyperparameters for this baseline model are as follows: \\
(top\_k = 5, num\_kernels = 6, embed = 'timeF', freq = 'h', d\_model = 32, d\_ff = 32, dropout = 0.1, encoder\_layers = 2) for the Weather, ETTh2, ETTm2, and EnergyData datasets, \\
(top\_k = 5, num\_kernels = 6, embed = 'timeF', freq = 'h', d\_model = 512, d\_ff = 512, dropout = 0.1, encoder\_layers = 2) for the Traffic dataset, \\
(top\_k = 5, num\_kernels = 6, embed = 'timeF', freq = 'h', d\_model = 256, d\_ff = 512, dropout = 0.1, encoder\_layers = 2) for the ECL and PEMS03 dataset, \\
(top\_k = 5, num\_kernels = 6, embed = 'timeF', freq = 'h', d\_model = 16, d\_ff = 32, dropout = 0.1, encoder\_layers = 2) for the ETTh1 and ETTm1 datasets, \\
(top\_k = 5, num\_kernels = 6, embed = 'timeF', freq = 'h', d\_model = 64, d\_ff = 64, dropout = 0.1, encoder\_layers = 2) for the Exchange dataset. \\
The BSA module is implemented at the beginning part of the forward pass right after normalization from the Non-stationary Transformer~\cite{Stationary} and right before the input embedding. The BSA module is implemented for each channel, which is the number of input variables for this case.

6. \textbf{FreTS}~\cite{FreTS}: The hyperparameters for this baseline model are as follows: \\
(embed\_size = 128, hidden\_size = 256, sparsity\_threshold = 0.01, scale = 0.02) for all datasets. Based on the paper, the channel-independent strategy is selected. \\
The BSA module is implemented at the very beginning of the forward pass, right before the token embedding. The BSA module is implemented for each channel, which is the number of input variables for this case.

7. \textbf{RLinear}~\cite{RLinear}: The only hyperparameter for this baseline model is the (dropout = 0.1). We set the Individual to True so that there are separate linear models for each number of input variables.
The BSA module is implemented at the beginning part of the forward pass right after normalization from the RevIN~\cite{RevIN} and right before the linear layer. The BSA module is implemented for each channel, which is the number of input variables for this case.

\subsection{Details on Training}
\label{appen:A.3}
\textbf{Pre-training and Finetuning configurations.}
To show how our BSA module performs when added to the baseline models, we saturated the models by greedy hyperparameter search.

The hyperparameter search space for the base model is as follows: the possible learning rate is (0.03, 0.01, 0.003, 0.001, 0.0003), and the weight decay is (0.01, 0.003, 0.001, 0.0003, 0.0001, 0.00003). 

The hyperparameter search space for BSA finetuning is as follows: the possible learning rate for the SA-Matrix in the BSA module is (0.08, 0.05, 0.03, 0.01, 0.003, 0.001), learning rate for the rest of the model, i.e. original modules, is (0.01, 0.003, 0.001, 0.0003, 0.0001, 0.00003, 0.00001), learning rate for smoothing factor $\alpha_{k}$ is (none, 0.03, 0.01, 0.003, 0.001, 0.0001, 0.00001), initialization for smoothing factor $\alpha_{k}$ is ([0.9, 0.99, 0.999], [0.9, 0.99, 0.999, 0.999], [0.9, 0.95, 0.992, 0.999], [0.8, 0.96, 0.992, 0.9984, 0.99968]).

\textbf{Model selection} 
Hyperparameter search is conducted based on the validation set. While models trained using conventional sample shuffling evenly represent the entire time series distribution from which the dataset is sampled, BSA learns data in chronological order. Consequently, the final model tends to favor the distribution of the later part of the data. This can be seen as a mild version of catastrophic forgetting, commonly occurring in continual learning. To mitigate this effect, we assigned higher weights to the later samples during the validation process. The weights are represented by $0.5+0.5\times \sin{(\frac{\pi}{2}\times\frac{val\_idx}{val\_len})}$, continuously increasing from 0.5 for the first sample to 1.0 for the last sample.

\textbf{Optimization}
The whole code is implemented in PyTorch~\cite{PyTorch}. Each experiment was conducted on a single NVIDIA GeForce RTX 3090Ti or NVIDIA A40 or NVIDIA L40 GPU. 
The default batch size for baseline model saturation is 64, while for our method—which involves fine-tuning after integrating the BSA module—it is 256. If a baseline model is too heavy and results in GPU memory overflow, the batch size is adjusted to fit within the available memory.
We used the ADAM~\cite{Adam} optimizer and L2 loss (MSE loss) for the model optimization. The baseline saturation training epoch is set to 40 epochs for the Weather dataset, 20 epochs for the Traffic dataset, and 30 epochs for the rest of the datasets. The finetuning epoch is set to 30 epochs for the ETTh1, ETTh2, and Exchange datasets and 20 epochs for the rest of the datasets.

\section{Real world dataset experiments} 
\label{appen:result1}
\subsection{Full experiment results}
\label{appen:result1-1}
The experiments were conducted on a total of 11 public real-world datasets and 7 forecasting models. The average results (MSE, MAE) of the experiments conducted with three random seeds 0, 1, and 2 are shown in Table \ref{tab:full_table}, and the standard deviations are shown in Table \ref{tab:full_std}. The values reported in Table \ref{tab:short_main} are labeled as Avg in Table \ref{tab:full_table}. For the Exchange dataset, we only report experiments with a prediction length of \{96\}. Using prediction lengths of \{192, 336, 720\} causes improper best model selection due to an insufficient number of validation samples. The empty results indicate that the training is too heavy, and the results are not yet available. We expect to have the results ready by the rebuttal. 

\begin{sidewaystable}
\caption{Full average results of three random seeds for prediction lengths $S \in \{96, 192, 336, 720\}$ with a fixed look-back window $T = \{96\}$. The average values across all prediction lengths are reported in Table \ref{tab:short_main}. The Exchange dataset is only done with prediction lengths of $S \in \{96\}$ due to its short data length. The setting for the Illness dataset is $T = \{36\}$, $S \in \{24, 36, 48, 60\}$.}
\label{tab:full_table}
\resizebox{\textwidth}{!}{%
\begin{tabular}{@{}cc|cccc|cccc|cccc|cccc|cccc|cccc|cccc@{}}
\toprule
\multicolumn{2}{c|}{\multirow{3}{*}{}} & \multicolumn{4}{c|}{Dlinear} & \multicolumn{4}{c|}{RLinear} & \multicolumn{4}{c|}{FreTS} & \multicolumn{4}{c|}{TimesNet} & \multicolumn{4}{c|}{iTransformer} & \multicolumn{4}{c|}{Crossformer} & \multicolumn{4}{c}{PatchTST} \\ \cmidrule(l){3-30} 
\multicolumn{2}{c|}{} & \multicolumn{2}{c|}{base} & \multicolumn{2}{c|}{BSA} & \multicolumn{2}{c|}{base} & \multicolumn{2}{c|}{BSA} & \multicolumn{2}{c|}{base} & \multicolumn{2}{c|}{BSA} & \multicolumn{2}{c|}{base} & \multicolumn{2}{c|}{BSA} & \multicolumn{2}{c|}{base} & \multicolumn{2}{c|}{BSA} & \multicolumn{2}{c|}{base} & \multicolumn{2}{c|}{BSA} & \multicolumn{2}{c|}{base} & \multicolumn{2}{c}{BSA} \\ \cmidrule(l){3-30} 
\multicolumn{2}{c|}{} & MSE & \multicolumn{1}{c|}{MAE} & MSE & MAE & MSE & \multicolumn{1}{c|}{MAE} & MSE & MAE & MSE & \multicolumn{1}{c|}{MAE} & MSE & MAE & MSE & \multicolumn{1}{c|}{MAE} & MSE & MAE & MSE & \multicolumn{1}{c|}{MAE} & MSE & MAE & MSE & \multicolumn{1}{c|}{MAE} & MSE & MAE & MSE & \multicolumn{1}{c|}{MAE} & MSE & MAE \\ \midrule
\multicolumn{1}{c|}{\multirow{5}{*}{Weather}} & 96 & 0.1630 & \multicolumn{1}{c|}{0.2363} & 0.1519 & 0.2203 & 0.1656 & \multicolumn{1}{c|}{0.2108} & 0.1594 & 0.2065 & 0.1670 & \multicolumn{1}{c|}{0.2235} & 0.1587 & 0.2214 & 0.1673 & \multicolumn{1}{c|}{0.2157} & 0.1647 & 0.2135 & 0.1706 & \multicolumn{1}{c|}{0.2096} & 0.1598 & 0.2037 & 0.1475 & \multicolumn{1}{c|}{0.2151} & 0.1447 & 0.2140 & 0.1742 & \multicolumn{1}{c|}{0.2149} & 0.1633 & 0.2096 \\
\bottomrule
\end{tabular}%
}
\end{sidewaystable}
\begin{sidewaystable}
\caption{Full standard deviation results of three random seeds with prediction lengths $S \in \{96, 192, 336, 720\}$ and a fixed lookback length $T = \{96\}$. The Exchange dataset is only done with prediction lengths of $S \in \{96\}$ due to its short data length. The setting for the Illness dataset is $T = \{36\}$, $S \in \{24, 36, 48, 60\}$.}
\label{tab:full_std}
\resizebox{\textwidth}{!}{%
\begin{tabular}{@{}cc|cccc|cccc|cccc|cccc|cccc|cccc|cccc@{}}
\toprule
\multicolumn{2}{c|}{\multirow{3}{*}{}} & \multicolumn{4}{c|}{Dlinear} & \multicolumn{4}{c|}{RLinear} & \multicolumn{4}{c|}{FreTS} & \multicolumn{4}{c|}{TimesNet} & \multicolumn{4}{c|}{iTransformer} & \multicolumn{4}{c|}{Crossformer} & \multicolumn{4}{c}{PatchTST} \\ \cmidrule(l){3-30} 
\multicolumn{2}{c|}{} & \multicolumn{2}{c|}{base} & \multicolumn{2}{c|}{BSA} & \multicolumn{2}{c|}{base} & \multicolumn{2}{c|}{BSA} & \multicolumn{2}{c|}{base} & \multicolumn{2}{c|}{BSA} & \multicolumn{2}{c|}{base} & \multicolumn{2}{c|}{BSA} & \multicolumn{2}{c|}{base} & \multicolumn{2}{c|}{BSA} & \multicolumn{2}{c|}{base} & \multicolumn{2}{c|}{BSA} & \multicolumn{2}{c|}{base} & \multicolumn{2}{c}{BSA} \\ \cmidrule(l){3-30} 
\multicolumn{2}{c|}{} & MSE & \multicolumn{1}{c|}{MAE} & MSE & MAE & MSE & \multicolumn{1}{c|}{MAE} & MSE & MAE & MSE & \multicolumn{1}{c|}{MAE} & MSE & MAE & MSE & \multicolumn{1}{c|}{MAE} & MSE & MAE & MSE & \multicolumn{1}{c|}{MAE} & MSE & MAE & MSE & \multicolumn{1}{c|}{MAE} & MSE & MAE & MSE & \multicolumn{1}{c|}{MAE} & MSE & MAE \\ \midrule
\multicolumn{1}{c|}{\multirow{5}{*}{Weather}} & 96 & 0.0002 & \multicolumn{1}{c|}{0.0005} & 0.0000 & 0.0000 & 0.0002 & \multicolumn{1}{c|}{0.0003} & 0.0000 & 0.0000 & 0.0001 & \multicolumn{1}{c|}{0.0002} & 0.0003 & 0.0005 & 0.0008 & \multicolumn{1}{c|}{0.0002} & 0.0015 & 0.0011 & 0.0015 & \multicolumn{1}{c|}{0.0017} & 0.0008 & 0.0011 & 0.0024 & \multicolumn{1}{c|}{0.0012} & 0.0029 & 0.0027 & 0.0004 & \multicolumn{1}{c|}{0.0005} & 0.0012 & 0.0013 \\
\bottomrule
\end{tabular}%
}
\end{sidewaystable}

\subsection{SA-Matrix and FFT visualization}
\label{appen:result1-2}
\begin{figure}[h]
        \centering
        \includegraphics[width=\linewidth]{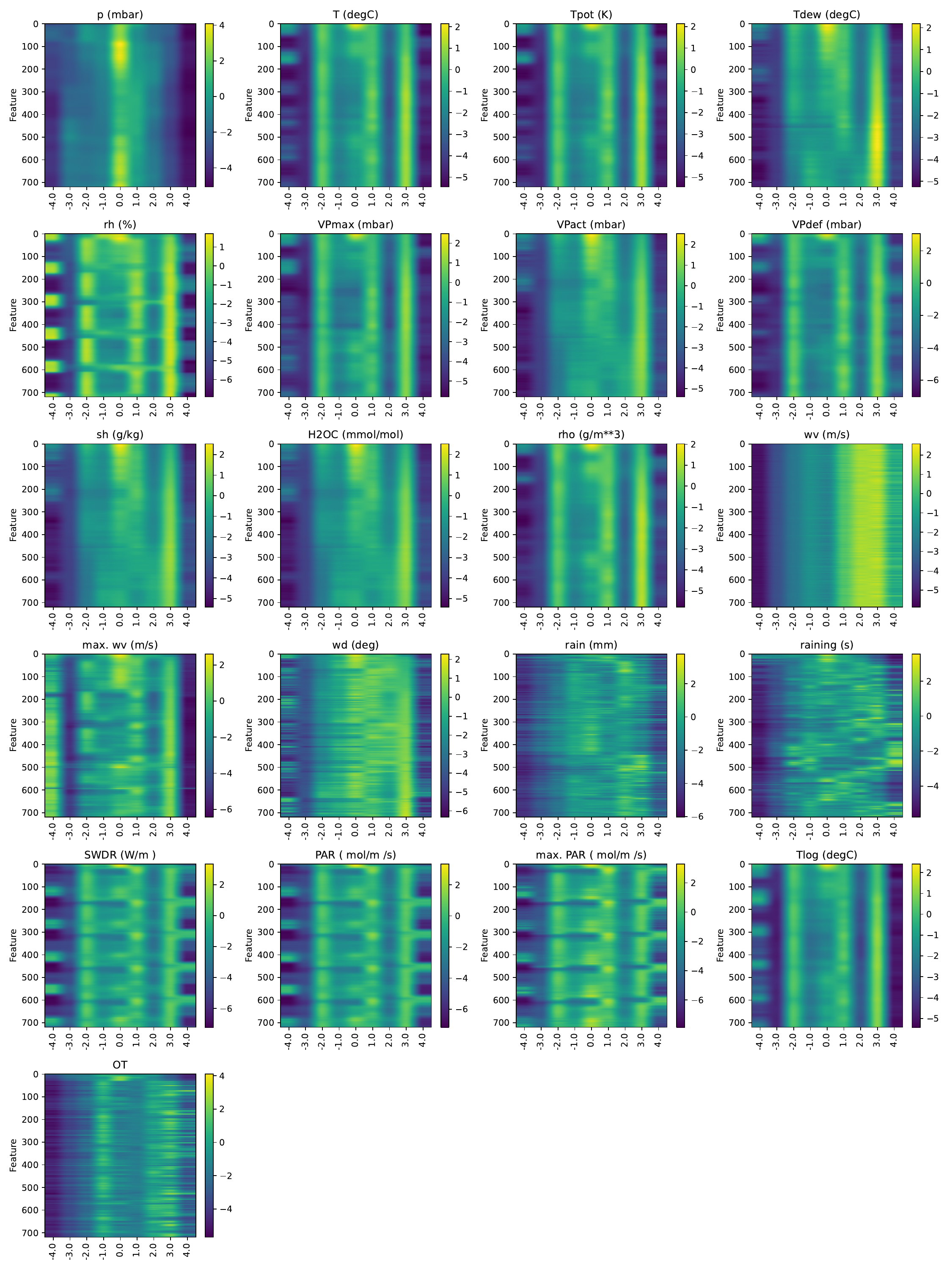}
        \caption{The 2D heatmaps of the SA-Matrix from the DLinear model finetuned with a 720 prediction length on the Weather dataset.}
        \label{fig:heatmap}
    \end{figure}

\begin{figure}[h]
        \centering
        \includegraphics[width=\linewidth]{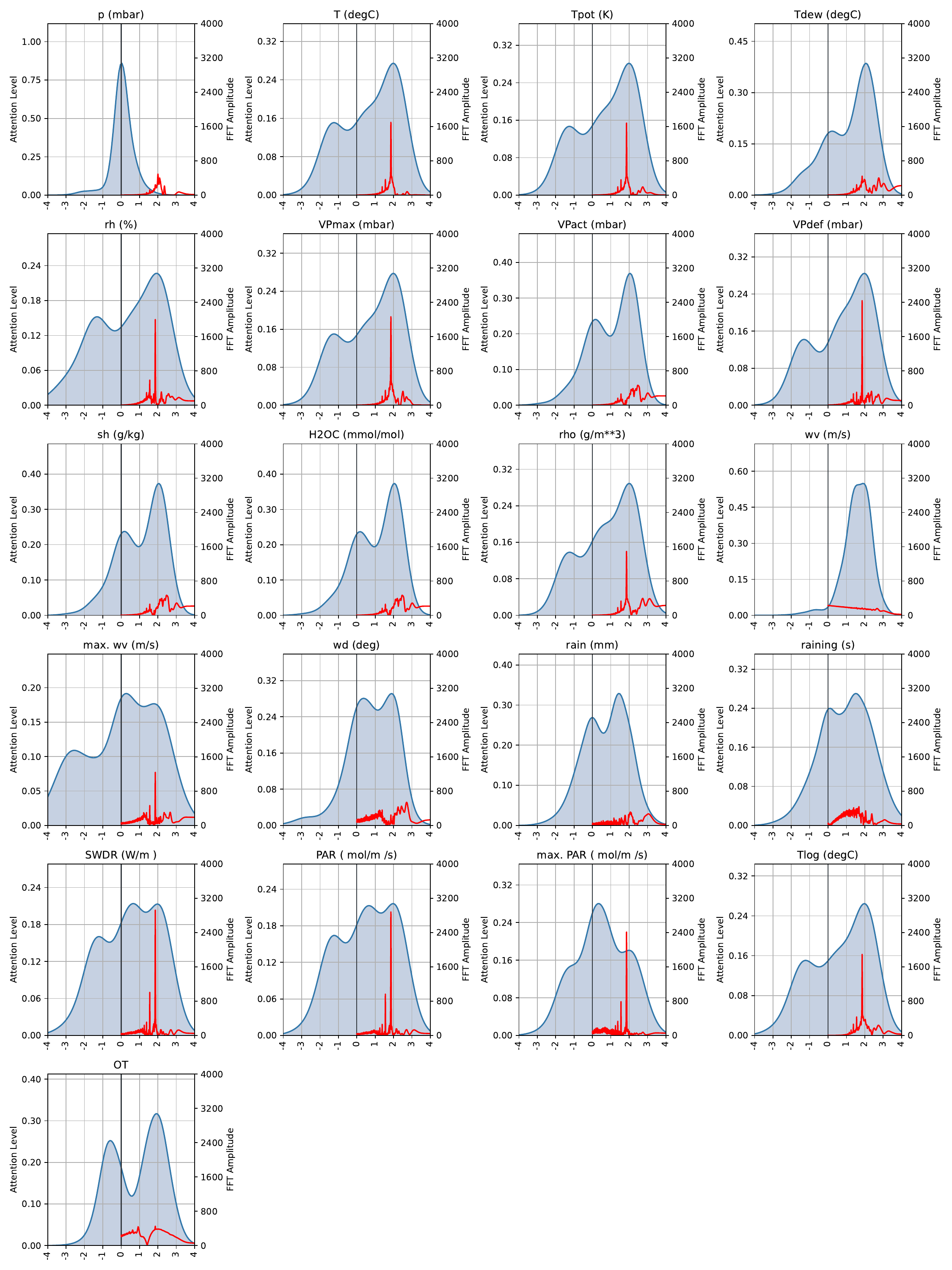}
        \caption{The kernel density estimate graphs of the SA-Matrix from the DLinear model finetuned with a 720 prediction length on the Weather dataset and FFT graphs of the data.}
        \label{fig:DLinear weather kde}
    \end{figure}

\begin{figure}[h]
        \centering
        \includegraphics[width=\linewidth]{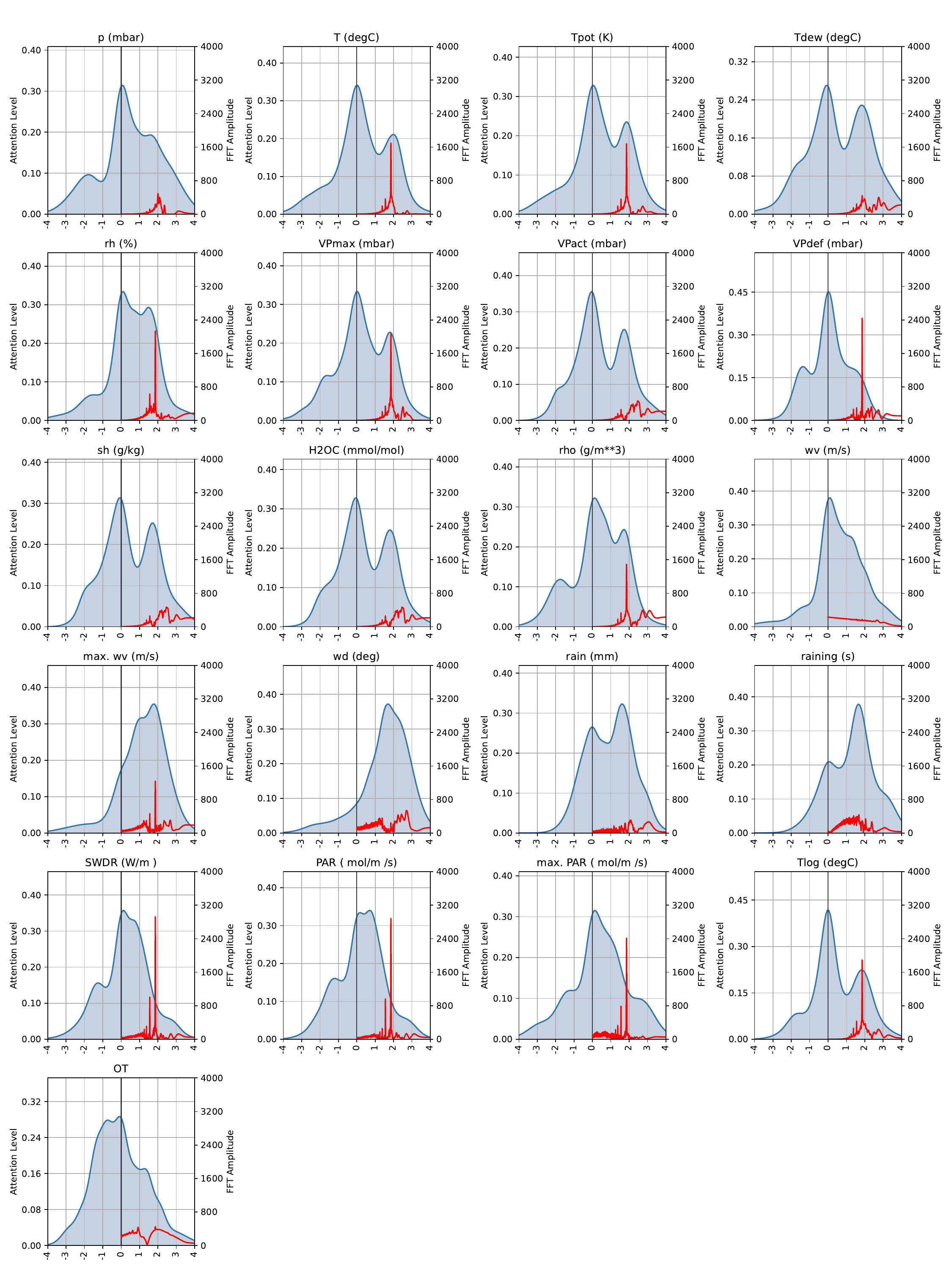}
        \caption{The kernel density estimate graphs of the SA-Matrix from the iTransformer model finetuned with a 720 prediction length on the Weather Data and FFT graphs of the data.}
        \label{fig:iTransformer weather kde}
    \end{figure}

\begin{figure}[h]
        \centering
        \includegraphics[width=\linewidth]{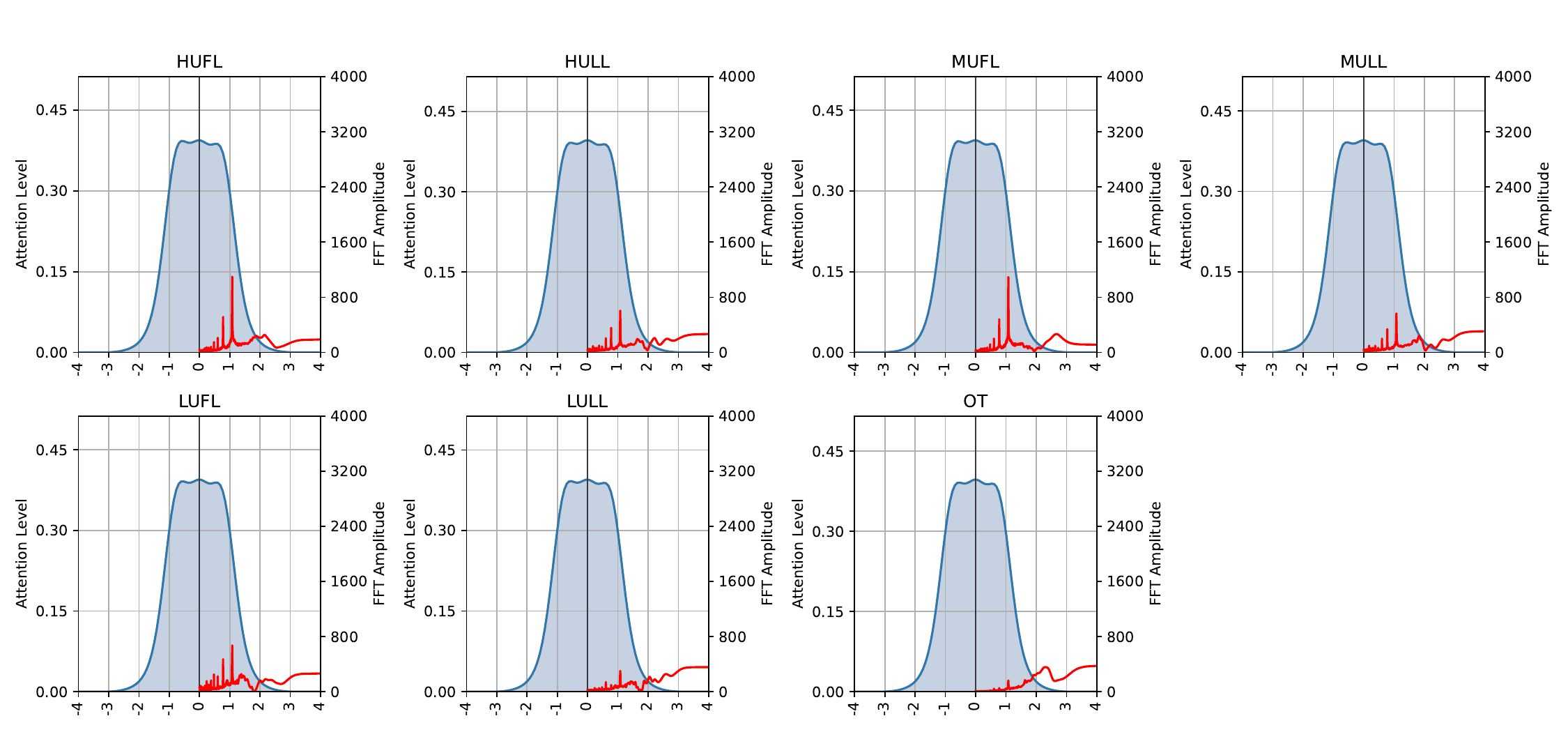}
        \caption{The kernel density estimate graphs of the SA-Matrix from the DLinear model finetuned with a 720 prediction length on the ETTh1 dataset and FFT graphs of the data.}
        \label{fig:ETTh1 kde}
    \end{figure}

\begin{figure}[h]
        \centering
        \includegraphics[width=\linewidth]{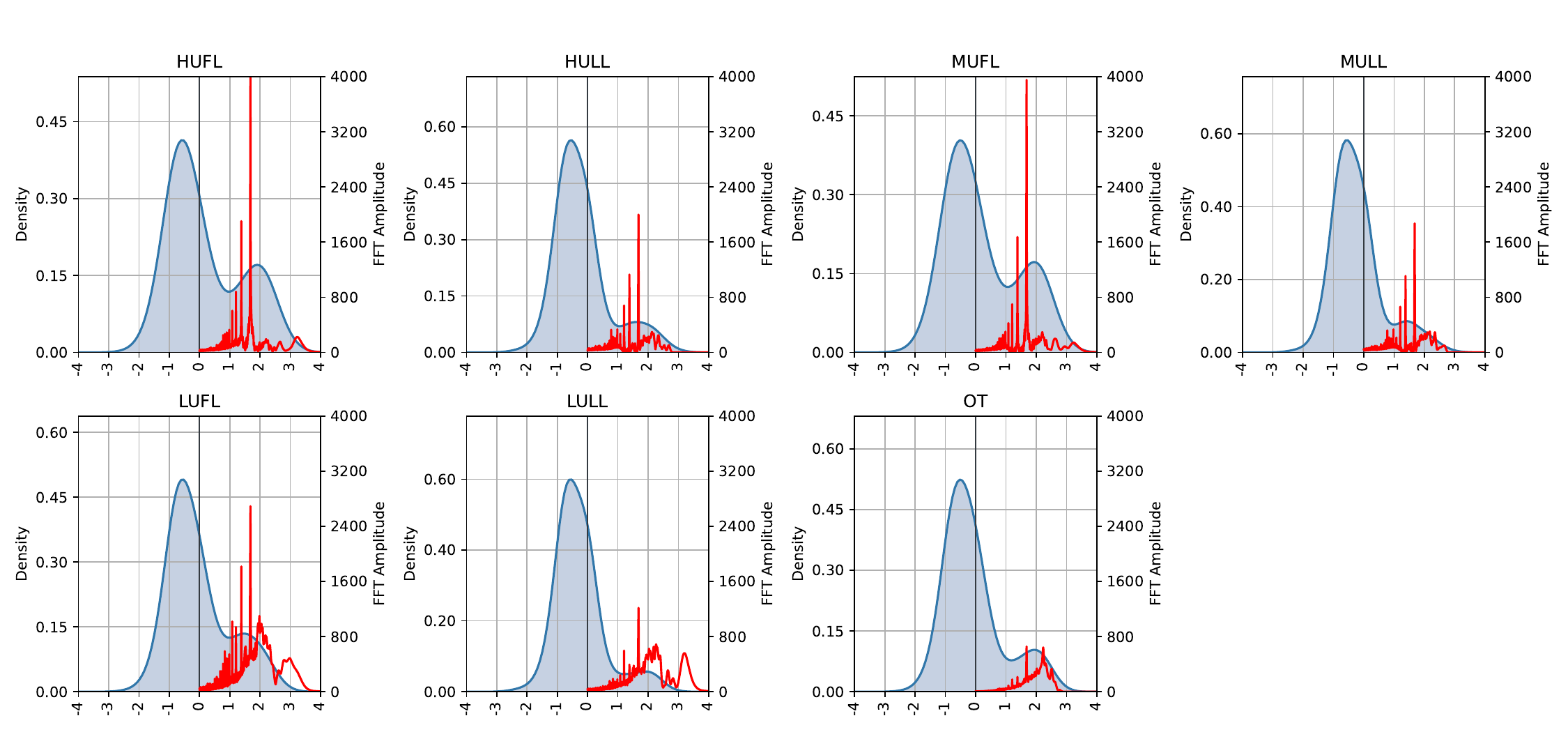}
        \caption{The kernel density estimate graphs of the SA-Matrix from the DLinear model finetuned with a 720 prediction length on the ETTm1 dataset and FFT graphs of the data.}
        \label{fig:ETTm1 kde}
    \end{figure}

\begin{figure}[!t]
    \centering
    \includegraphics[width=\linewidth]{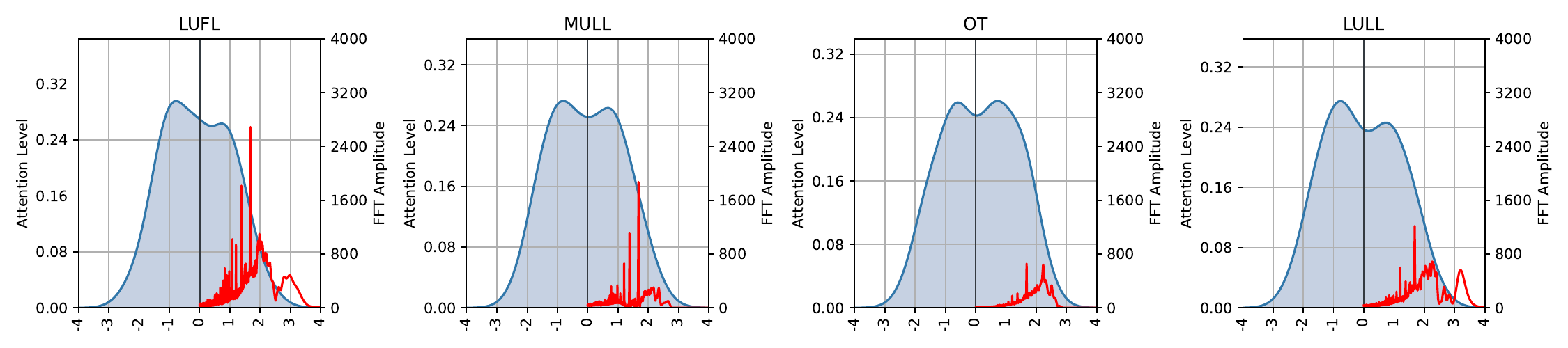}
    \caption{The kernel density estimate graphs of the SA-Matrix from the iTransformer model finetuned with a 720 prediction length on the ETTm1 dataset and FFT graphs of the data.}
    \label{fig:ETTm1 iTransformer}
\end{figure}

To investigate how the SA-Matrix of BSA predominantly attends to specific frequency bands, we employed the heatmaps, the Gaussian kernel density estimate graphs, and the FFT graphs as shown in Figure~\ref{fig:heatmap}, ~\ref{fig:DLinear weather kde}, ~\ref{fig:iTransformer weather kde}, ~\ref{fig:ETTh1 kde}, ~\ref{fig:ETTm1 iTransformer} and ~\ref{fig:ETTm1 kde}.

The 2D heatmap depicts the learnable parameters of the SA-Matrix, defined as  \textit{sa-matrix}$\in \mathbb{R}^{(2K+1) \times D}$, where $K$ represents the number of smoothing factors and $D$ represents the dimension of the input features. Consequently, the y-axis of the 2D heatmap matches the length of $D$, and the x-axis corresponds to $2K+1$, symmetrically arranged around zero. On the x-axis, positive values indicate the low-frequency regions, while negative values represent the high-frequency regions.

The Gaussian kernel density estimate graphs intuitively reveal which frequency bands the SA-Matrix predominantly attends to. The $K$ smoothing factors were modified according to Equation ~\ref{eq:8} and symmetrically arranged around zero, serving as the data points for kernel density estimation (KDE). The weight values from each row of the SA-Matrix were converted to probabilities using the softmax function, and the resulting outputs established a mapping of these weight values to data points necessary for Gaussian KDE. This mapping facilitated the construction of the Gaussian KDE graph. Subsequently, the overall probability density of the SA-Matrix was estimated by calculating the mean across columns. Consequently, the y-axis of the graphs denotes the attention level of the SA-Matrix, whereas the x-axis indicates the scalar indices, encompassing the range of the smoothing factors. To better represent the variation in weight values across frequency bands, we adjusted the bandwidth of the KDE function to 0.4, based on the product of the Gaussian kernel’s covariance factor and the standard deviation of the sampled weights in the matrix.

\begin{gather}
\label{eq:8}
    \alpha^{'}_{k} = \log_{10}(\dfrac{1}{1-\sigma(\alpha_{k})})
\end{gather}


The FFT graph is depicted with a red line. FFT analysis was conducted for each variable to determine if the fine-tuned SA-Matrix aligns with the frequencies exhibited by the dataset. To compare the low-frequency parts, the negative log-scale was used on the x-axis, showing that progression to the right indicates decreasing frequencies. The Gaussian filter with a fixed standard deviation of 5 was utilized to smooth the signal’s amplitude.

\begin{figure}[t]
    \centering
    \begin{subfigure}[b]{0.325\columnwidth}
        \centering
        \includegraphics[width=\linewidth]{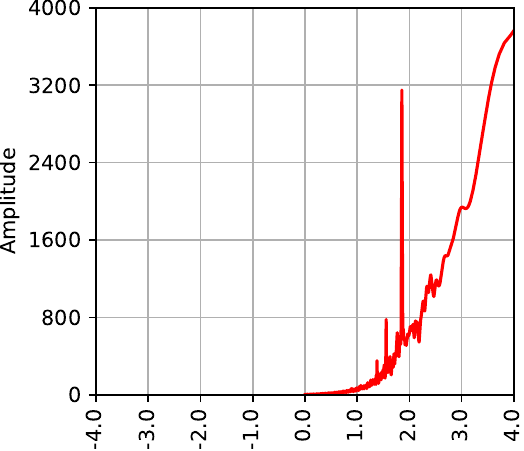}
        \caption{Pre-Denoising}
    \end{subfigure}
    \begin{subfigure}[b]{0.325\columnwidth}
        \centering
        \includegraphics[width=\linewidth]{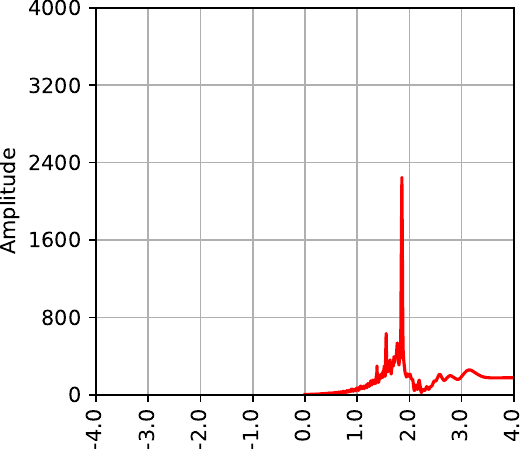}
        \caption{Post-denoising}
    \end{subfigure}
    \begin{subfigure}[b]{0.28\columnwidth}
        \centering
        \includegraphics[width=\linewidth]{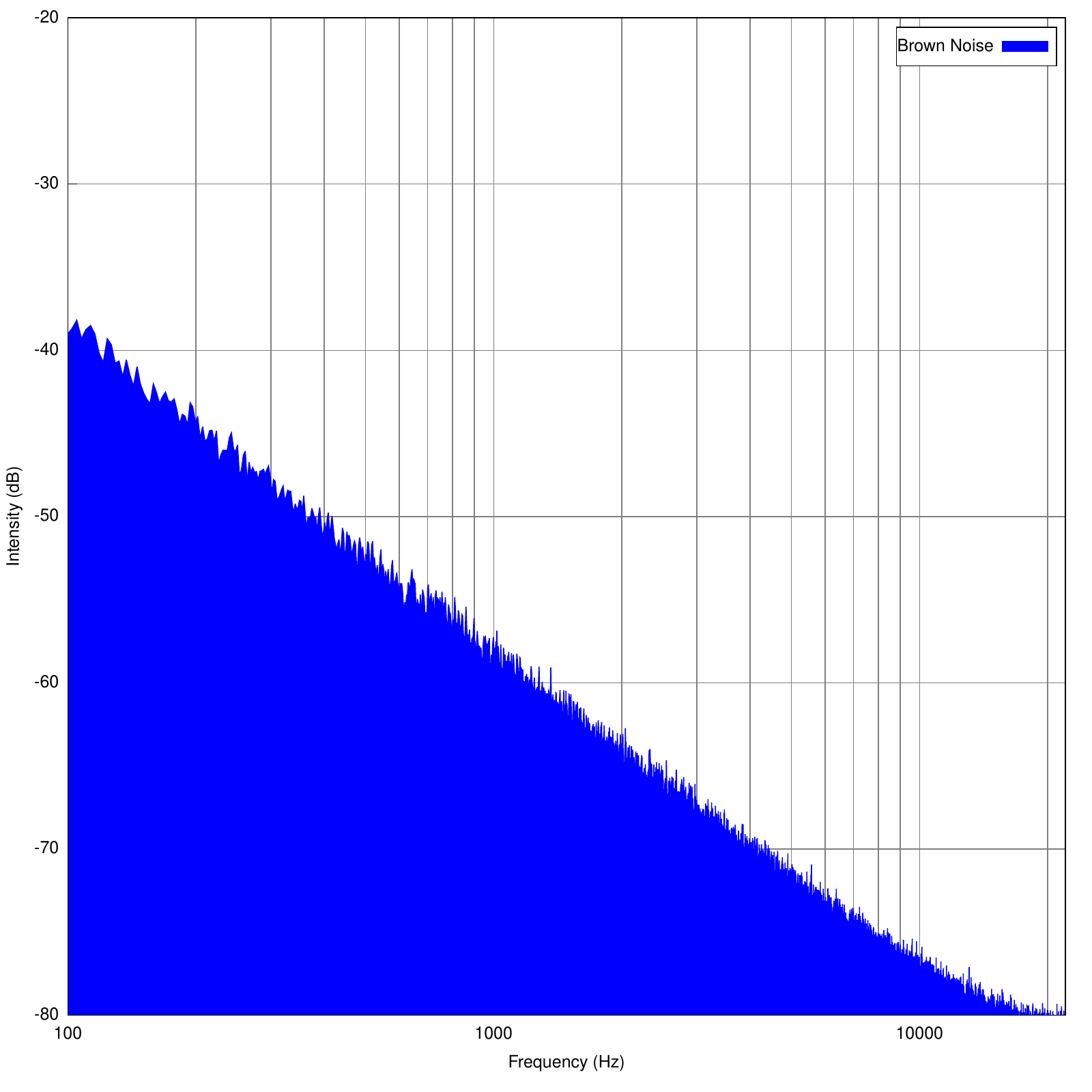}
        \caption{Brownian Noise~\cite{wiki:BrownianNoise}}
    \end{subfigure}
        \caption {The illustration compares the pre-denoising and post-denoising results alongside Brownian noise. For (a) and (b), the y-axis represents amplitude, and the x-axis is on a negative log scale, indicating that moving toward the right corresponds to lower frequencies. Conversely, in (c), the y-axis represents decibels, and the x-axis is on a log scale, where moving toward the left corresponds to lower frequencies.}
        \label{fig:denoising phase}
\end{figure}

The denoising phase was executed on the FFT output to analyze the correlation between the frequency tendencies of the dataset and the manipulated weights in the SA-Matrix. Figure~\ref{fig:denoising phase}(c) depicts Brownian noise, with the x-axis denoting frequency and the y-axis in decibels (dB). Both axes are configured on a log scale. Converting the y-axis from decibels to amplitude is akin to implementing linear scaling, which reveals a steeper rise in noise at lower frequencies. This noise generally arises in the low-frequency bands of most real-world systems, such as electronic systems and environmental sciences~\cite{krapf2018power}. Figure~\ref{fig:denoising phase} (a) shows that noise escalates within the low-frequency regions across variables, leading to the assumption that unique Brownian noise originates in each system. Consequently, a linear approximation was used for the denoising function, which was performed manually. The outcomes of this process are presented in Figure~\ref{fig:denoising phase} (b).

\begin{figure}[!t]
    \centering
    \includegraphics[width=\linewidth]{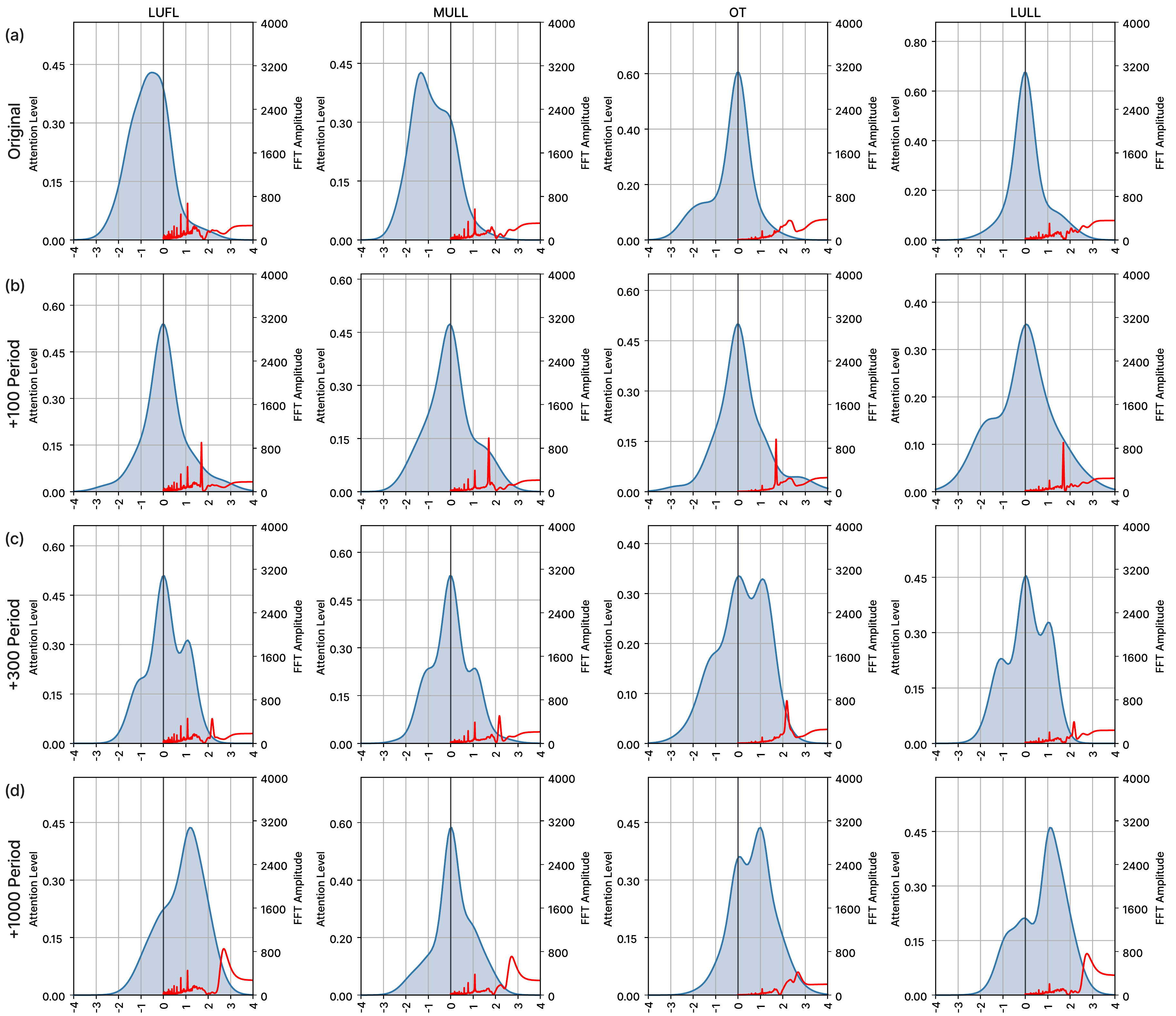}
    \caption{The kernel density estimate graphs of LUFL, MULL, OT, and LULL channels of the SA-Matrix from the iTransformer model with a 720 prediction length on the original and synthetic ETTh1 data. Row (a) represents the original data, while rows (b) - (d) display synthetic datasets. These datasets were generated by adding sine waves with periods of 100, 300, and 1000, respectively.}
    \label{fig:synthethic4x4}
\end{figure}

\begin{figure}[!t]
    \centering
    \includegraphics[width=\linewidth]{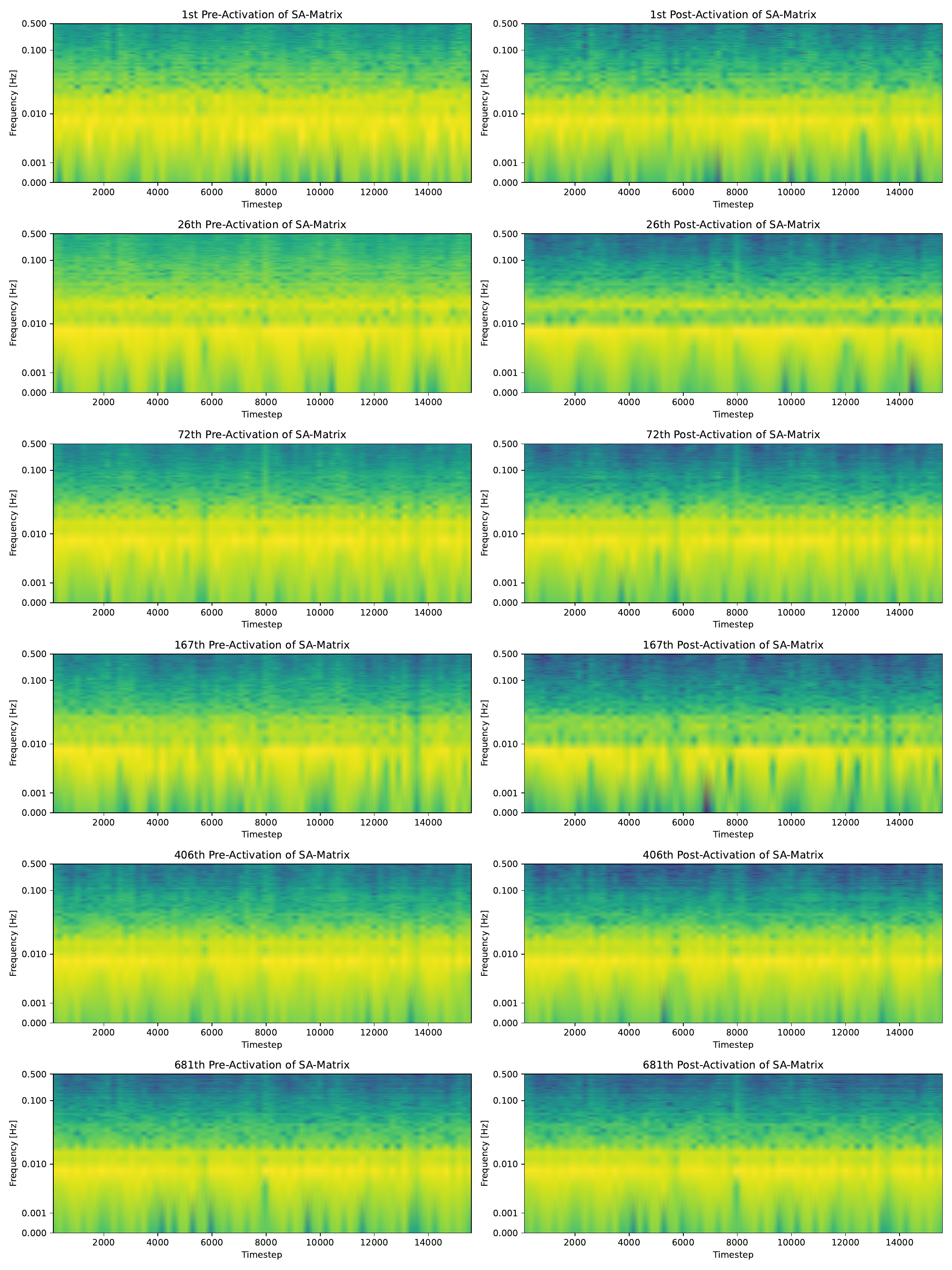}
    \caption{Illustration of the SWDR (W/m) channel of the Pre-Activation (left) and Post-Activation (right) through SA-Matrix from the DLinear model trained with a 720 prediction length on the weather dataset.}
    \label{fig:spectrogram}
\end{figure}

\section{Synthetic dataset experiments}
\label{appen:result2}
We constructed synthetic datasets by adding sine waves with periods of 100, 300, and 1000 to each channel (variable) of the ETTh1 and ETTh2 datasets. 
The scale of the sine wave is set to the standard deviation value of each channel.
The phases of the sine waves for each channel are randomly sampled from a uniform distribution from 0 to $2\pi$ to make decorrelated sine waves. The full results are reported in Table \ref{tab:synthetic}. 

\begin{table}
\caption{Full average results of three random seeds with prediction lengths $S \in \{96, 192, 336, 720\}$ and a fixed lookback length $T = \{96\}$ on synthetic data.  The original results are from the public datasets ETTh1 and ETTh2, as shown in Table \ref{tab:full_table}. Synthetic datasets were created by adding sine waves with periods of 100, 300, and 1000 to the original ETTh datasets. Gains mean performance improvements by using our method.}
\vspace{5pt}
\label{tab:synthetic}
\resizebox{\textwidth}{!}{%
\begin{tabular}{@{}cr|cccccc|cc|cccccc@{}}
\toprule
\multicolumn{2}{l|}{\multirow{3}{*}{}} & \multicolumn{6}{c|}{iTransformer} & \multicolumn{2}{l|}{\multirow{3}{*}{}} & \multicolumn{6}{c}{iTransformer} \\
\multicolumn{2}{l|}{} & \multicolumn{2}{c}{base} & \multicolumn{2}{c}{BSA} & \multicolumn{2}{c|}{Gain(\%)} & \multicolumn{2}{l|}{} & \multicolumn{2}{c}{base} & \multicolumn{2}{c}{BSA} & \multicolumn{2}{c}{Gain(\%)} \\
\multicolumn{2}{l|}{} & MSE & MAE & MSE & MAE & MSE & MAE & \multicolumn{2}{l|}{} & MSE & MAE & MSE & MAE & MSE & MAE \\ \midrule
\multirow{5}{*}{\begin{tabular}[c]{@{}c@{}}ETTh1\\ original\end{tabular}} & 96 & 0.4316 & 0.4453 & 0.4277 & 0.4426 & 0.909 & 0.605 & \multirow{5}{*}{\begin{tabular}[c]{@{}c@{}}ETTh2\\ original\end{tabular}} & 96 & 0.2397 & 0.3278 & 0.2347 & 0.3236 & 2.062 & 1.260 \\
 & 192 & 0.4906 & 0.4857 & 0.4809 & 0.4812 & 1.973 & 0.931 &  & 192 & 0.2950 & 0.3641 & 0.2901 & 0.3624 & 1.654 & 0.467 \\
 & 336 & 0.5451 & 0.5217 & 0.5401 & 0.5243 & 0.908 & -0.504 &  & 336 & 0.3316 & 0.3880 & 0.3271 & 0.3884 & 1.370 & -0.103 \\
 & 720 & 0.6991 & 0.6174 & 0.6979 & 0.6204 & 0.172 & -0.493 &  & 720 & 0.4160 & 0.4389 & 0.4144 & 0.4391 & 0.364 & -0.040 \\
 & avg & 0.5416 & 0.5175 & 0.5367 & 0.5171 & 0.991 & 0.135 &  & avg & 0.3206 & 0.3797 & 0.3166 & 0.3784 & 1.362 & 0.396 \\ \midrule
\multirow{5}{*}{\begin{tabular}[c]{@{}c@{}}ETTh1\\ 100\end{tabular}} & 96 & 0.3108 & 0.3889 & 0.2980 & 0.3800 & 4.111 & 2.286 & \multirow{5}{*}{\begin{tabular}[c]{@{}c@{}}ETTh2\\ 100\end{tabular}} & 96 & 0.1608 & 0.2738 & 0.1565 & 0.2724 & 2.716 & 0.530 \\
 & 192 & 0.3470 & 0.4172 & 0.3298 & 0.4062 & 4.968 & 2.635 &  & 192 & 0.1880 & 0.2993 & 0.1839 & 0.3009 & 2.198 & -0.516 \\
 & 336 & 0.3825 & 0.4454 & 0.3719 & 0.4409 & 2.774 & 1.001 &  & 336 & 0.2098 & 0.3156 & 0.2070 & 0.3153 & 1.335 & 0.105 \\
 & 720 & 0.4825 & 0.5162 & 0.4784 & 0.5150 & 0.844 & 0.230 &  & 720 & 0.2580 & 0.3526 & 0.2558 & 0.3522 & 0.855 & 0.137 \\
 & avg & 0.3807 & 0.4419 & 0.3695 & 0.4355 & 3.174 & 1.538 &  & avg & 0.2042 & 0.3103 & 0.2008 & 0.3102 & 1.776 & 0.064 \\ \midrule
\multirow{5}{*}{\begin{tabular}[c]{@{}c@{}}ETTh1\\ 300\end{tabular}} & 96 & 0.5350 & 0.5291 & 0.3674 & 0.4334 & 31.330 & 18.081 & \multirow{5}{*}{\begin{tabular}[c]{@{}c@{}}ETTh2\\ 300\end{tabular}} & 96 & 0.2646 & 0.3663 & 0.1711 & 0.2957 & 35.319 & 19.272 \\
 & 192 & 0.6278 & 0.5873 & 0.4156 & 0.4729 & 33.799 & 19.476 &  & 192 & 0.3219 & 0.4107 & 0.2189 & 0.3356 & 32.010 & 18.290 \\
 & 336 & 0.5741 & 0.5608 & 0.4217 & 0.4780 & 26.537 & 14.763 &  & 336 & 0.2971 & 0.3921 & 0.2287 & 0.3444 & 23.023 & 12.178 \\
 & 720 & 0.7036 & 0.6361 & 0.5273 & 0.5514 & 25.068 & 13.313 &  & 720 & 0.3669 & 0.4410 & 0.2804 & 0.3837 & 23.582 & 12.999 \\
 & avg & 0.6101 & 0.5783 & 0.4330 & 0.4839 & 29.183 & 16.408 &  & avg & 0.3126 & 0.4025 & 0.2248 & 0.3398 & 28.483 & 15.685 \\ \midrule
\multirow{5}{*}{\begin{tabular}[c]{@{}c@{}}ETTh1\\ 1000\end{tabular}} & 96 & 0.3370 & 0.4082 & 0.3295 & 0.4036 & 2.225 & 1.135 & \multirow{5}{*}{\begin{tabular}[c]{@{}c@{}}ETTh2\\ 1000\end{tabular}} & 96 & 0.1880 & 0.3003 & 0.1799 & 0.2927 & 4.305 & 2.529 \\
 & 192 & 0.5019 & 0.5197 & 0.4392 & 0.4833 & 12.487 & 6.992 &  & 192 & 0.3283 & 0.4108 & 0.2742 & 0.3745 & 16.489 & 8.842 \\
 & 336 & 0.8177 & 0.6785 & 0.5441 & 0.5591 & 33.455 & 17.591 &  & 336 & 0.6335 & 0.5958 & 0.4320 & 0.4856 & 31.820 & 18.492 \\
 & 720 & 1.2900 & 0.8721 & 0.6741 & 0.6411 & 47.742 & 26.484 &  & 720 & 1.1044 & 0.7933 & 0.4554 & 0.5046 & 58.760 & 36.389 \\
 & avg & 0.7366 & 0.6196 & 0.4967 & 0.5218 & 23.978 & 13.050 &  & avg & 0.5636 & 0.5251 & 0.3354 & 0.4144 & 27.843 & 16.563 \\ \bottomrule
\end{tabular}%
}
\end{table}

\subsection{SA-Matrix and FFT visualization}
\label{appen:result3}

As illustrated in Figure~\ref{fig:synthethic4x4}, the LUFL, MULL, OT, and LULL channels are organized into columns, with each graph positioned within rows that correspond to the periods added to the dataset. Figure~\ref{fig:synthethic4x4} (a) shows the SA-Matrix from the iTransformer model fine-tuned with a 720 prediction length on the ETTh1 data that does not include added sine waves. From (b) to (d), the models were trained with synthetic datasets increasingly augmented with sine waves at periods of 100, 300, and 1000. As the number of added periods increases, there is a shift in the SA-Matrix weight distribution toward low-frequency patterns in each channel.

\section{Analysis and Ablation Studies}
\label{appen:result4}

\subsection{BSA insertion site analysis}
\label{appen:result4.1}
\begin{table}
\caption{Full results for different BSA module locations for iTransformer~\cite{iTransformer} and DLinear~\cite{DLinear} on the Weather dataset. Positions 1 and 4 are the only available options for the DLinear Model.}
\vspace{6pt}
\label{tab:different_location}
\resizebox{\textwidth}{!}{%
\begin{tabular}{@{}cr|cccccccccccccccc@{}}
\toprule
\multicolumn{2}{c|}{\multirow{3}{*}{}} & \multicolumn{16}{c}{iTransformer} \\
\multicolumn{2}{c|}{} & \multicolumn{2}{c}{base} & \multicolumn{2}{c}{1} & \multicolumn{2}{c}{2} & \multicolumn{2}{c}{3} & \multicolumn{2}{c}{4} & \multicolumn{2}{c}{5} & \multicolumn{2}{c}{6} & \multicolumn{2}{c}{7} \\
\multicolumn{2}{c|}{} & MSE & MAE & MSE & MAE & MSE & MAE & MSE & MAE & MSE & MAE & MSE & MAE & MSE & MAE & MSE & MAE \\ \midrule
\multirow{5}{*}{weather} & 96 & \multicolumn{1}{r}{0.1706} & \multicolumn{1}{r}{0.2096} & \multicolumn{1}{r}{0.1598} & \multicolumn{1}{r}{0.2037} & \multicolumn{1}{r}{0.1589} & \multicolumn{1}{r}{0.2027} & \multicolumn{1}{r}{0.1538} & \multicolumn{1}{r}{0.1990} & \multicolumn{1}{r}{0.1667} & \multicolumn{1}{r}{0.2077} & \multicolumn{1}{r}{0.1545} & \multicolumn{1}{r}{0.1999} & \multicolumn{1}{r}{0.1676} & \multicolumn{1}{r}{0.2080} & \multicolumn{1}{r}{0.1582} & \multicolumn{1}{r}{0.2011} \\
 & 192 & \multicolumn{1}{r}{0.2203} & \multicolumn{1}{r}{0.2544} & \multicolumn{1}{r}{0.2052} & \multicolumn{1}{r}{0.2466} & \multicolumn{1}{r}{0.2034} & \multicolumn{1}{r}{0.2463} & \multicolumn{1}{r}{0.1996} & \multicolumn{1}{r}{0.2441} & \multicolumn{1}{r}{0.2149} & \multicolumn{1}{r}{0.2526} & \multicolumn{1}{r}{0.2003} & \multicolumn{1}{r}{0.2451} & \multicolumn{1}{r}{0.2206} & \multicolumn{1}{r}{0.2548} & \multicolumn{1}{r}{0.2110} & \multicolumn{1}{r}{0.2504} \\
 & 336 & \multicolumn{1}{r}{0.2762} & \multicolumn{1}{r}{0.2952} & \multicolumn{1}{r}{0.2526} & \multicolumn{1}{r}{0.2856} & \multicolumn{1}{r}{0.2510} & \multicolumn{1}{r}{0.2846} & \multicolumn{1}{r}{0.2528} & \multicolumn{1}{r}{0.2837} & \multicolumn{1}{r}{0.2719} & \multicolumn{1}{r}{0.2940} & \multicolumn{1}{r}{0.2519} & \multicolumn{1}{r}{0.2855} & \multicolumn{1}{r}{0.2756} & \multicolumn{1}{r}{0.2948} & \multicolumn{1}{r}{0.2534} & \multicolumn{1}{r}{0.2854} \\
 & 720 & \multicolumn{1}{r}{0.3551} & \multicolumn{1}{r}{0.3473} & \multicolumn{1}{r}{0.3252} & \multicolumn{1}{r}{0.3371} & \multicolumn{1}{r}{0.3277} & \multicolumn{1}{r}{0.3386} & \multicolumn{1}{r}{0.3255} & \multicolumn{1}{r}{0.3349} & \multicolumn{1}{r}{0.3497} & \multicolumn{1}{r}{0.3465} & \multicolumn{1}{r}{0.3238} & \multicolumn{1}{r}{0.3380} & \multicolumn{1}{r}{0.3513} & \multicolumn{1}{r}{0.3472} & \multicolumn{1}{r}{0.3434} & \multicolumn{1}{r}{0.3440} \\
 & Avg & \multicolumn{1}{r}{0.2556} & \multicolumn{1}{r}{0.2766} & \multicolumn{1}{r}{0.2357} & \multicolumn{1}{r}{0.2682} & \multicolumn{1}{r}{0.2352} & \multicolumn{1}{r}{0.2681} & \multicolumn{1}{r}{0.2329} & \multicolumn{1}{r}{0.2654} & \multicolumn{1}{r}{0.2508} & \multicolumn{1}{r}{0.2752} & \multicolumn{1}{r}{0.2326} & \multicolumn{1}{r}{0.2671} & \multicolumn{1}{r}{0.2538} & \multicolumn{1}{r}{0.2762} & \multicolumn{1}{r}{0.2415} & \multicolumn{1}{r}{0.2702} \\ \midrule \midrule
\multicolumn{2}{l|}{\multirow{3}{*}{}} & \multicolumn{16}{c}{Dlinear} \\
\multicolumn{2}{l|}{} & \multicolumn{2}{c}{base} & \multicolumn{2}{c}{1} & \multicolumn{2}{c}{2} & \multicolumn{2}{c}{3} & \multicolumn{2}{c}{4} & \multicolumn{2}{c}{5} & \multicolumn{2}{c}{6} & \multicolumn{2}{c}{7} \\
\multicolumn{2}{l|}{} & MSE & MAE & MSE & MAE & MSE & MAE & MSE & MAE & MSE & MAE & MSE & MAE & MSE & MAE & MSE & MAE \\ \midrule
\multirow{5}{*}{weather} & 96 & 0.1630 & 0.2363 & 0.1565 & 0.2306 & - & - & - & - & 0.1519 & 0.2203 & - & - & - & - & - & - \\
 & 192 & 0.2091 & 0.2832 & 0.1985 & 0.2724 & - & - & - & - & 0.1912 & 0.2601 & - & - & - & - & - & - \\
 & 336 & 0.2634 & 0.3271 & 0.2590 & 0.3251 & - & - & - & - & 0.2375 & 0.2993 & - & - & - & - & - & - \\
 & 720 & 0.3421 & 0.3868 & 0.3167 & 0.3697 & - & - & - & - & 0.2979 & 0.3463 & - & - & - & - & - & - \\
 & Avg & 0.2444 & 0.3084 & 0.2327 & 0.2994 & - & - & - & - & 0.2196 & 0.2815 & - & - & - & - & - & - \\\bottomrule
\end{tabular}%
}
\end{table}
Table~\ref{tab:different_location} shows the full results for different BSA module insertion sites in the iTransformer~\cite{iTransformer} and DLinear~\cite{DLinear} models on the Weather Dataset.
Table \ref{tab:4.3.1} is from the average value of Table~\ref{tab:different_location}. The BSA module insert position can be found in Figure~\ref{fig:4.3.1}. 

\subsection{BSA variable input lengths analysis}
\label{appen:result4.2}
\begin{table}[!h]
\caption{Full result for the performance of the base model and BSA according to changes in Input length. Experiments were conducted on Weather and PEMS03 data using DLinear, RLinear, and iTransformer models.}
\vspace{6pt}
\label{tab:variable_input}
\resizebox{\textwidth}{!}{%
\begin{tabular}{cr|rrrr|rrrr|rrrr}
\toprule
\multicolumn{2}{l|}{\multirow{3}{*}{}} &
  \multicolumn{4}{c|}{Dlinear} &
  \multicolumn{4}{c|}{RLinear} &
  \multicolumn{4}{c}{iTransformer} \\
\multicolumn{2}{l|}{} &
  \multicolumn{2}{c}{base} &
  \multicolumn{2}{c|}{BSA} &
  \multicolumn{2}{c}{base} &
  \multicolumn{2}{c|}{BSA} &
  \multicolumn{2}{c}{base} &
  \multicolumn{2}{c}{BSA} \\
\multicolumn{2}{l|}{} &
  \multicolumn{1}{c}{MSE} &
  \multicolumn{1}{c}{MAE} &
  \multicolumn{1}{c}{MSE} &
  \multicolumn{1}{c|}{MAE} &
  \multicolumn{1}{c}{MSE} &
  \multicolumn{1}{c}{MAE} &
  \multicolumn{1}{c}{MSE} &
  \multicolumn{1}{c|}{MAE} &
  \multicolumn{1}{c}{MSE} &
  \multicolumn{1}{c}{MAE} &
  \multicolumn{1}{c}{MSE} &
  \multicolumn{1}{c}{MAE} \\ 
  \midrule
  \midrule
\multirow{5}{*}{\begin{tabular}[c]{@{}c@{}}Weather\\ 48\end{tabular}} &
  96 &
  0.2187 &
  0.2950 &
  0.1616 &
  0.2332 &
  0.1898 &
  0.2322 &
  0.1783 &
  0.2233 &
  0.2054 &
  0.2271 &
  0.1713 &
  0.2124 \\
 &
  192 &
  0.2477 &
  0.3203 &
  0.1994 &
  0.2715 &
  0.2315 &
  0.2654 &
  0.2206 &
  0.2583 &
  0.2490 &
  0.2695 &
  0.2178 &
  0.2541 \\
 &
  336 &
  0.3086 &
  0.3696 &
  0.2460 &
  0.3094 &
  0.2897 &
  0.3064 &
  0.2768 &
  0.2980 &
  0.3075 &
  0.3099 &
  0.2716 &
  0.2947 \\
 &
  720 &
  0.3817 &
  0.4204 &
  0.3085 &
  0.3585 &
  0.3699 &
  0.3566 &
  0.3579 &
  0.3499 &
  0.3831 &
  0.3595 &
  0.3459 &
  0.3442 \\
 &
  Avg &
  0.2892 &
  0.3513 &
  0.2289 &
  0.2932 &
  0.2702 &
  0.2902 &
  0.2584 &
  0.2824 &
  0.2862 &
  0.2915 &
  0.2517 &
  0.2763 \\ \midrule
\multirow{5}{*}{\begin{tabular}[c]{@{}c@{}}Weather\\ 96\end{tabular}} &
  96 &
  0.1630 &
  0.2363 &
  0.1519 &
  0.2203 &
  0.1656 &
  0.2108 &
  0.1594 &
  0.2065 &
  0.1706 &
  0.2096 &
  0.1598 &
  0.2037 \\
 &
  192 &
  0.2091 &
  0.2832 &
  0.1912 &
  0.2601 &
  0.2119 &
  0.2512 &
  0.2013 &
  0.2444 &
  0.2203 &
  0.2544 &
  0.2052 &
  0.2466 \\
 &
  336 &
  0.2634 &
  0.3271 &
  0.2375 &
  0.2993 &
  0.2679 &
  0.2911 &
  0.2541 &
  0.2823 &
  0.2762 &
  0.2952 &
  0.2526 &
  0.2856 \\
 &
  720 &
  0.3421 &
  0.3868 &
  0.2979 &
  0.3463 &
  0.3465 &
  0.3412 &
  0.3315 &
  0.3338 &
  0.3551 &
  0.3473 &
  0.3252 &
  0.3371 \\
 &
  Avg &
  0.2444 &
  0.3084 &
  0.2196 &
  0.2815 &
  0.2480 &
  0.2736 &
  0.2366 &
  0.2667 &
  0.2556 &
  0.2766 &
  0.2357 &
  0.2682 \\ \midrule
\multirow{5}{*}{\begin{tabular}[c]{@{}c@{}}Weather\\ 192\end{tabular}} &
  96 &
  0.1514 &
  0.2208 &
  0.1466 &
  0.2138 &
  0.1522 &
  0.1982 &
  0.1471 &
  0.1960 &
  0.1650 &
  0.2100 &
  0.1521 &
  0.2008 \\
 &
  192 &
  0.1952 &
  0.2660 &
  0.1856 &
  0.2530 &
  0.1963 &
  0.2383 &
  0.1898 &
  0.2360 &
  0.2098 &
  0.2510 &
  0.2023 &
  0.2450 \\
 &
  336 &
  0.2492 &
  0.3100 &
  0.2315 &
  0.2914 &
  0.2502 &
  0.2785 &
  0.2393 &
  0.2733 &
  0.2625 &
  0.2903 &
  0.2428 &
  0.2810 \\
 &
  720 &
  0.3324 &
  0.3771 &
  0.2958 &
  0.3436 &
  0.3285 &
  0.3313 &
  0.3111 &
  0.3242 &
  0.3356 &
  0.3410 &
  0.3089 &
  0.3291 \\
 &
  Avg &
  0.2321 &
  0.2934 &
  0.2149 &
  0.2754 &
  0.2318 &
  0.2616 &
  0.2218 &
  0.2574 &
  0.2432 &
  0.2731 &
  0.2265 &
  0.2640 \\ 
  \midrule
  \midrule
\multirow{5}{*}{\begin{tabular}[c]{@{}c@{}}PEMS03\\ 48\end{tabular}} &
  96 &
  0.5159 &
  0.5565 &
  0.4340 &
  0.5021 &
  1.2342 &
  0.8375 &
  1.0040 &
  0.7564 &
  0.4214 &
  0.4518 &
  0.2934 &
  0.3715 \\
 &
  192 &
  0.5673 &
  0.5921 &
  0.4646 &
  0.5287 &
  1.7909 &
  1.0790 &
  1.3785 &
  0.9286 &
  0.4010 &
  0.4454 &
  0.2551 &
  0.3513 \\
 &
  336 &
  0.5056 &
  0.5437 &
  0.4018 &
  0.4767 &
  1.2898 &
  0.8454 &
  1.0455 &
  0.7490 &
  0.4196 &
  0.4485 &
  0.3084 &
  0.3801 \\
 &
  720 &
  0.5524 &
  0.5811 &
  0.4398 &
  0.5047 &
  1.5496 &
  0.9620 &
  1.3303 &
  0.8777 &
  0.4699 &
  0.4848 &
  0.3507 &
  0.4100 \\
 &
  Avg &
  0.5353 &
  0.5684 &
  0.4351 &
  0.5031 &
  1.4661 &
  0.9310 &
  1.1896 &
  0.8279 &
  0.4280 &
  0.4576 &
  0.3019 &
  0.3782 \\ \midrule
\multirow{5}{*}{\begin{tabular}[c]{@{}c@{}}PEMS03\\ 96\end{tabular}} &
  96 &
  0.4405 &
  0.5097 &
  0.4121 &
  0.4875 &
  1.0366 &
  0.7815 &
  0.6789 &
  0.6359 &
  0.2213 &
  0.3226 &
  0.1569 &
  0.2689 \\
 &
  192 &
  0.4658 &
  0.5259 &
  0.3876 &
  0.4677 &
  1.0995 &
  0.7937 &
  0.7128 &
  0.6326 &
  0.2687 &
  0.3566 &
  0.1865 &
  0.2949 \\
 &
  336 &
  0.3973 &
  0.4669 &
  0.3461 &
  0.4317 &
  0.8227 &
  0.6434 &
  0.5861 &
  0.5408 &
  0.2479 &
  0.3303 &
  0.1979 &
  0.2957 \\
 &
  720 &
  0.4420 &
  0.5048 &
  0.3922 &
  0.4703 &
  1.0101 &
  0.7432 &
  0.7194 &
  0.6184 &
  0.3093 &
  0.3716 &
  0.2488 &
  0.3330 \\
 &
  Avg &
  0.4364 &
  0.5018 &
  0.3845 &
  0.4643 &
  0.9922 &
  0.7405 &
  0.6743 &
  0.6069 &
  0.2618 &
  0.3453 &
  0.1975 &
  0.2981 \\ \midrule
\multirow{5}{*}{\begin{tabular}[c]{@{}c@{}}PEMS03\\ 192\end{tabular}} &
  96 &
  0.3904 &
  0.4570 &
  0.2660 &
  0.3698 &
  0.4078 &
  0.4618 &
  0.2674 &
  0.3609 &
  0.1463 &
  0.2569 &
  0.1218 &
  0.2323 \\
 &
  192 &
  0.3156 &
  0.3932 &
  0.2543 &
  0.3552 &
  0.3290 &
  0.3916 &
  0.2581 &
  0.3394 &
  0.1637 &
  0.2641 &
  0.1450 &
  0.2484 \\
 &
  336 &
  0.3078 &
  0.3903 &
  0.2650 &
  0.3641 &
  0.3276 &
  0.3890 &
  0.2723 &
  0.3495 &
  0.1738 &
  0.2680 &
  0.1623 &
  0.2596 \\
 &
  720 &
  0.3588 &
  0.4370 &
  0.3221 &
  0.4110 &
  0.3991 &
  0.4418 &
  0.3467 &
  0.4038 &
  0.2226 &
  0.3000 &
  0.2091 &
  0.2924 \\
 &
  Avg &
  0.3432 &
  0.4194 &
  0.2768 &
  0.3750 &
  0.3659 &
  0.4211 &
  0.2861 &
  0.3634 &
  0.1766 &
  0.2723 &
  0.1596 &
  0.2582\\
  \bottomrule
  
\end{tabular}
}
\end{table}
Table~\ref{tab:variable_input} shows the full results of the performance of the base model and BSA according to changes in Input length. Experiments were conducted on Weather and PEMS03 data using DLinear~\cite{DLinear}, RLinear~\cite{RLinear}, and iTransformer~\cite{iTransformer} models. Table~\ref{tab:input_length} in the main text shows the summarized results.

\subsection{BSA computational cost analysis}
\label{appen:result4.3}
\begin{table}[ht]
\caption{Full result for the model training time (sec/1 step), peak memory usage (GB), and the number of parameters (M) for both the base model and the model with BSA applied. The experiments were conducted using the lightweight Weather dataset with 21 channels and the heavy PEMS03 dataset with 358 channels. Tests were performed on Timesnet, iTransformer, Crossformer, and PatchTST models.}
\vspace{6pt}
\label{tab:computational_cost}
\resizebox{\textwidth}{!}{%
\begin{tabular}{cc|ccc|ccc|ccc|ccc}
\toprule
\multicolumn{2}{c|}{} &
  \multicolumn{3}{c|}{TimesNet} &
  \multicolumn{3}{c|}{iTransformer} &
  \multicolumn{3}{c|}{Crossformer} &
  \multicolumn{3}{c}{PatchTST} \\
\rowcolor[HTML]{96FFFB}
\midrule
\multicolumn{2}{c|}{\cellcolor[HTML]{96FFFB}Time (sec/1step)} &
  base &
  BSA &
  gain(\%) &
  base &
  BSA &
  \cellcolor[HTML]{96FFFB}gain(\%) &
  base &
  BSA &
  \cellcolor[HTML]{96FFFB}gain(\%) &
  base &
  BSA &
  \cellcolor[HTML]{96FFFB}gain(\%) \\ \cline{3-14} 
 &
  \cellcolor[HTML]{EFEFEF}96 &
  \cellcolor[HTML]{EFEFEF}0.066 &
  \cellcolor[HTML]{EFEFEF}0.070 &
  \cellcolor[HTML]{EFEFEF}5.37 &
  \cellcolor[HTML]{EFEFEF}0.024 &
  \cellcolor[HTML]{EFEFEF}0.029 &
  \cellcolor[HTML]{EFEFEF}20.71 &
  \cellcolor[HTML]{EFEFEF}0.076 &
  \cellcolor[HTML]{EFEFEF}0.079 &
  \cellcolor[HTML]{EFEFEF}3.58 &
  \cellcolor[HTML]{EFEFEF}0.045 &
  \cellcolor[HTML]{EFEFEF}0.046 &
  \cellcolor[HTML]{EFEFEF}1.52 \\
 &
  192 &
  0.076 &
  0.072 &
  -4.56 &
  0.024 &
  0.025 &
  3.92 &
  0.072 &
  0.078 &
  8.28 &
  0.046 &
  0.048 &
  3.01 \\
 &
  \cellcolor[HTML]{EFEFEF}336 &
  \cellcolor[HTML]{EFEFEF}0.081 &
  \cellcolor[HTML]{EFEFEF}0.078 &
  \cellcolor[HTML]{EFEFEF}-3.30 &
  \cellcolor[HTML]{EFEFEF}0.025 &
  \cellcolor[HTML]{EFEFEF}0.027 &
  \cellcolor[HTML]{EFEFEF}8.05 &
  \cellcolor[HTML]{EFEFEF}0.077 &
  \cellcolor[HTML]{EFEFEF}0.080 &
  \cellcolor[HTML]{EFEFEF}3.66 &
  \cellcolor[HTML]{EFEFEF}0.050 &
  \cellcolor[HTML]{EFEFEF}0.052 &
  \cellcolor[HTML]{EFEFEF}5.12 \\
 &
  720 &
  0.104 &
  0.107 &
  2.47 &
  0.025 &
  0.033 &
  30.74 &
  0.075 &
  0.080 &
  6.50 &
  0.048 &
  0.049 &
  2.88 \\
\multirow{-5}{*}{Weather} &
  \cellcolor[HTML]{EFEFEF}Avg &
  \cellcolor[HTML]{EFEFEF}0.082 &
  \cellcolor[HTML]{EFEFEF}0.082 &
  \cellcolor[HTML]{EFEFEF}0.00 &
  \cellcolor[HTML]{EFEFEF}0.024 &
  \cellcolor[HTML]{EFEFEF}0.028 &
  \cellcolor[HTML]{EFEFEF}15.86 &
  \cellcolor[HTML]{EFEFEF}0.075 &
  \cellcolor[HTML]{EFEFEF}0.079 &
  \cellcolor[HTML]{EFEFEF}5.50 &
  \cellcolor[HTML]{EFEFEF}0.047 &
  \cellcolor[HTML]{EFEFEF}0.049 &
  \cellcolor[HTML]{EFEFEF}3.13 \\ \hline
 &
  96 &
  1.160 &
  1.147 &
  -1.06 &
  0.077 &
  0.080 &
  3.96 &
  0.163 &
  0.162 &
  -0.95 &
  0.878 &
  0.877 &
  -0.10 \\
 &
  \cellcolor[HTML]{EFEFEF}192 &
  \cellcolor[HTML]{EFEFEF}1.791 &
  \cellcolor[HTML]{EFEFEF}1.717 &
  \cellcolor[HTML]{EFEFEF}-4.12 &
  \cellcolor[HTML]{EFEFEF}0.079 &
  \cellcolor[HTML]{EFEFEF}0.082 &
  \cellcolor[HTML]{EFEFEF}3.69 &
  \cellcolor[HTML]{EFEFEF}0.296 &
  \cellcolor[HTML]{EFEFEF}0.290 &
  \cellcolor[HTML]{EFEFEF}-1.98 &
  \cellcolor[HTML]{EFEFEF}0.889 &
  \cellcolor[HTML]{EFEFEF}0.882 &
  \cellcolor[HTML]{EFEFEF}-0.74 \\
 &
  336 &
  2.374 &
  2.502 &
  5.40 &
  0.097 &
  0.096 &
  -1.26 &
  0.492 &
  0.485 &
  -1.45 &
  0.881 &
  0.883 &
  0.27 \\
 &
  \cellcolor[HTML]{EFEFEF}720 &
  \cellcolor[HTML]{EFEFEF}4.397 &
  \cellcolor[HTML]{EFEFEF}4.425 &
  \cellcolor[HTML]{EFEFEF}0.63 &
  \cellcolor[HTML]{EFEFEF}0.115 &
  \cellcolor[HTML]{EFEFEF}0.118 &
  \cellcolor[HTML]{EFEFEF}2.57 &
  \cellcolor[HTML]{EFEFEF}0.965 &
  \cellcolor[HTML]{EFEFEF}0.929 &
  \cellcolor[HTML]{EFEFEF}-3.78 &
  \cellcolor[HTML]{EFEFEF}0.908 &
  \cellcolor[HTML]{EFEFEF}0.902 &
  \cellcolor[HTML]{EFEFEF}-0.57 \\
\multirow{-5}{*}{PEMS03} &
  Avg &
  2.430 &
  2.448 &
  0.21 &
  0.092 &
  0.094 &
  2.24 &
  0.479 &
  0.466 &
  -2.04 &
  0.889 &
  0.886 &
  -0.29 \\
\rowcolor[HTML]{96FFFB} 
\midrule
\multicolumn{2}{c|}{\cellcolor[HTML]{96FFFB}Memory (GB)} &
  base &
  BSA &
  \cellcolor[HTML]{96FFFB}gain(\%) &
  base &
  BSA &
  \cellcolor[HTML]{96FFFB}gain(\%) &
  base &
  BSA &
  \cellcolor[HTML]{96FFFB}gain(\%) &
  base &
  BSA &
  \cellcolor[HTML]{96FFFB}gain(\%) \\
 &
  96 &
  0.43 &
  0.43 &
  0.04 &
  0.20 &
  0.25 &
  24.65 &
  0.27 &
  0.28 &
  2.85 &
  1.59 &
  1.60 &
  0.36 \\
 &
  \cellcolor[HTML]{EFEFEF}192 &
  \cellcolor[HTML]{EFEFEF}0.57 &
  \cellcolor[HTML]{EFEFEF}0.58 &
  \cellcolor[HTML]{EFEFEF}1.18 &
  \cellcolor[HTML]{EFEFEF}0.21 &
  \cellcolor[HTML]{EFEFEF}0.32 &
  \cellcolor[HTML]{EFEFEF}54.08 &
  \cellcolor[HTML]{EFEFEF}0.48 &
  \cellcolor[HTML]{EFEFEF}0.49 &
  \cellcolor[HTML]{EFEFEF}1.11 &
  \cellcolor[HTML]{EFEFEF}1.60 &
  \cellcolor[HTML]{EFEFEF}1.61 &
  \cellcolor[HTML]{EFEFEF}0.47 \\
 &
  336 &
  0.79 &
  0.81 &
  2.15 &
  0.22 &
  0.29 &
  30.93 &
  0.87 &
  0.88 &
  0.81 &
  1.63 &
  1.63 &
  0.30 \\
 &
  \cellcolor[HTML]{EFEFEF}720 &
  \cellcolor[HTML]{EFEFEF}1.38 &
  \cellcolor[HTML]{EFEFEF}1.37 &
  \cellcolor[HTML]{EFEFEF}-0.07 &
  \cellcolor[HTML]{EFEFEF}0.25 &
  \cellcolor[HTML]{EFEFEF}0.30 &
  \cellcolor[HTML]{EFEFEF}19.57 &
  \cellcolor[HTML]{EFEFEF}2.28 &
  \cellcolor[HTML]{EFEFEF}2.28 &
  \cellcolor[HTML]{EFEFEF}0.24 &
  \cellcolor[HTML]{EFEFEF}1.68 &
  \cellcolor[HTML]{EFEFEF}1.68 &
  \cellcolor[HTML]{EFEFEF}0.25 \\
\multirow{-5}{*}{Weather} &
  Avg &
  0.79 &
  0.80 &
  0.82 &
  0.22 &
  0.29 &
  32.31 &
  0.97 &
  0.98 &
  1.25 &
  1.62 &
  1.63 &
  0.35 \\ \hline
 &
  \cellcolor[HTML]{EFEFEF}96 &
  \cellcolor[HTML]{EFEFEF}5.79 &
  \cellcolor[HTML]{EFEFEF}6.01 &
  \cellcolor[HTML]{EFEFEF}3.66 &
  \cellcolor[HTML]{EFEFEF}3.57 &
  \cellcolor[HTML]{EFEFEF}3.68 &
  \cellcolor[HTML]{EFEFEF}3.04 &
  \cellcolor[HTML]{EFEFEF}7.23 &
  \cellcolor[HTML]{EFEFEF}7.25 &
  \cellcolor[HTML]{EFEFEF}0.32 &
  \cellcolor[HTML]{EFEFEF}25.40 &
  \cellcolor[HTML]{EFEFEF}25.52 &
  \cellcolor[HTML]{EFEFEF}0.46 \\
 &
  192 &
  7.32 &
  7.59 &
  3.74 &
  3.65 &
  3.75 &
  2.76 &
  12.24 &
  12.26 &
  0.22 &
  25.49 &
  25.60 &
  0.44 \\
 &
  \cellcolor[HTML]{EFEFEF}336 &
  \cellcolor[HTML]{EFEFEF}10.26 &
  \cellcolor[HTML]{EFEFEF}10.01 &
  \cellcolor[HTML]{EFEFEF}-2.46 &
  \cellcolor[HTML]{EFEFEF}3.80 &
  \cellcolor[HTML]{EFEFEF}3.89 &
  \cellcolor[HTML]{EFEFEF}2.30 &
  \cellcolor[HTML]{EFEFEF}19.96 &
  \cellcolor[HTML]{EFEFEF}19.98 &
  \cellcolor[HTML]{EFEFEF}0.10 &
  \cellcolor[HTML]{EFEFEF}25.61 &
  \cellcolor[HTML]{EFEFEF}25.71 &
  \cellcolor[HTML]{EFEFEF}0.38 \\
 &
  720 &
  17.07 &
  16.46 &
  -3.58 &
  4.18 &
  4.23 &
  1.27 &
  42.34 &
  42.36 &
  0.05 &
  25.95 &
  26.02 &
  0.24 \\
\multirow{-5}{*}{PEMS03} &
  \cellcolor[HTML]{EFEFEF}Avg &
  \cellcolor[HTML]{EFEFEF}10.11 &
  \cellcolor[HTML]{EFEFEF}10.02 &
  \cellcolor[HTML]{EFEFEF}0.34 &
  \cellcolor[HTML]{EFEFEF}3.80 &
  \cellcolor[HTML]{EFEFEF}3.89 &
  \cellcolor[HTML]{EFEFEF}2.34 &
  \cellcolor[HTML]{EFEFEF}20.44 &
  \cellcolor[HTML]{EFEFEF}20.46 &
  \cellcolor[HTML]{EFEFEF}0.17 &
  \cellcolor[HTML]{EFEFEF}25.61 &
  \cellcolor[HTML]{EFEFEF}25.71 &
  \cellcolor[HTML]{EFEFEF}0.38 \\
\rowcolor[HTML]{96FFFB} 
\midrule
\multicolumn{2}{c|}{\cellcolor[HTML]{96FFFB}Parameters (M)} &
  base &
  BSA &
  \cellcolor[HTML]{96FFFB}gain(\%) &
  base &
  BSA &
  \cellcolor[HTML]{96FFFB}gain(\%) &
  base &
  BSA &
  \cellcolor[HTML]{96FFFB}gain(\%) &
  base &
  BSA &
  \cellcolor[HTML]{96FFFB}gain(\%) \\
 &
  \cellcolor[HTML]{EFEFEF}96 &
  \cellcolor[HTML]{EFEFEF}1.354 &
  \cellcolor[HTML]{EFEFEF}1.368 &
  \cellcolor[HTML]{EFEFEF}1.05 &
  \cellcolor[HTML]{EFEFEF}4.848 &
  \cellcolor[HTML]{EFEFEF}4.852 &
  \cellcolor[HTML]{EFEFEF}0.08 &
  \cellcolor[HTML]{EFEFEF}0.283 &
  \cellcolor[HTML]{EFEFEF}0.301 &
  \cellcolor[HTML]{EFEFEF}6.41 &
  \cellcolor[HTML]{EFEFEF}9.46 &
  \cellcolor[HTML]{EFEFEF}9.48 &
  \cellcolor[HTML]{EFEFEF}0.15 \\
 &
  192 &
  1.363 &
  1.381 &
  1.34 &
  4.897 &
  4.897 &
  0.00 &
  0.291 &
  0.313 &
  7.62 &
  10.05 &
  10.07 &
  0.18 \\
 &
  \cellcolor[HTML]{EFEFEF}336 &
  \cellcolor[HTML]{EFEFEF}1.377 &
  \cellcolor[HTML]{EFEFEF}1.391 &
  \cellcolor[HTML]{EFEFEF}1.03 &
  \cellcolor[HTML]{EFEFEF}4.971 &
  \cellcolor[HTML]{EFEFEF}4.971 &
  \cellcolor[HTML]{EFEFEF}0.00 &
  \cellcolor[HTML]{EFEFEF}0.302 &
  \cellcolor[HTML]{EFEFEF}0.317 &
  \cellcolor[HTML]{EFEFEF}4.67 &
  \cellcolor[HTML]{EFEFEF}10.94 &
  \cellcolor[HTML]{EFEFEF}10.96 &
  \cellcolor[HTML]{EFEFEF}0.17 \\
 &
  720 &
  1.414 &
  1.433 &
  1.29 &
  5.168 &
  5.172 &
  0.08 &
  0.333 &
  0.347 &
  4.24 &
  13.30 &
  13.32 &
  0.14 \\
\multirow{-5}{*}{Weather} &
  \cellcolor[HTML]{EFEFEF}Avg &
  \cellcolor[HTML]{EFEFEF}1.377 &
  \cellcolor[HTML]{EFEFEF}1.393 &
  \cellcolor[HTML]{EFEFEF}1.18 &
  \cellcolor[HTML]{EFEFEF}4.971 &
  \cellcolor[HTML]{EFEFEF}4.973 &
  \cellcolor[HTML]{EFEFEF}0.04 &
  \cellcolor[HTML]{EFEFEF}0.302 &
  \cellcolor[HTML]{EFEFEF}0.320 &
  \cellcolor[HTML]{EFEFEF}5.73 &
  \cellcolor[HTML]{EFEFEF}10.94 &
  \cellcolor[HTML]{EFEFEF}10.96 &
  \cellcolor[HTML]{EFEFEF}0.16 \\ \hline
 &
  96 &
  151.6 &
  151.9 &
  0.16 &
  4.834 &
  5.076 &
  5.00 &
  10.69 &
  10.94 &
  2.26 &
  9.46 &
  9.71 &
  2.55 \\
 &
  \cellcolor[HTML]{EFEFEF}192 &
  \cellcolor[HTML]{EFEFEF}151.6 &
  \cellcolor[HTML]{EFEFEF}151.9 &
  \cellcolor[HTML]{EFEFEF}0.16 &
  \cellcolor[HTML]{EFEFEF}4.883 &
  \cellcolor[HTML]{EFEFEF}5.125 &
  \cellcolor[HTML]{EFEFEF}4.95 &
  \cellcolor[HTML]{EFEFEF}11.45 &
  \cellcolor[HTML]{EFEFEF}11.69 &
  \cellcolor[HTML]{EFEFEF}2.11 &
  \cellcolor[HTML]{EFEFEF}10.05 &
  \cellcolor[HTML]{EFEFEF}10.30 &
  \cellcolor[HTML]{EFEFEF}2.40 \\
 &
  336 &
  151.6 &
  151.9 &
  0.16 &
  4.957 &
  5.199 &
  4.87 &
  12.57 &
  12.81 &
  1.92 &
  10.94 &
  11.18 &
  2.21 \\
 &
  \cellcolor[HTML]{EFEFEF}720 &
  \cellcolor[HTML]{EFEFEF}151.7 &
  \cellcolor[HTML]{EFEFEF}151.9 &
  \cellcolor[HTML]{EFEFEF}0.16 &
  \cellcolor[HTML]{EFEFEF}5.154 &
  \cellcolor[HTML]{EFEFEF}5.396 &
  \cellcolor[HTML]{EFEFEF}4.69 &
  \cellcolor[HTML]{EFEFEF}15.58 &
  \cellcolor[HTML]{EFEFEF}15.82 &
  \cellcolor[HTML]{EFEFEF}1.55 &
  \cellcolor[HTML]{EFEFEF}13.30 &
  \cellcolor[HTML]{EFEFEF}13.54 &
  \cellcolor[HTML]{EFEFEF}1.82 \\
\multirow{-5}{*}{PEMS03} &
  Avg &
  151.6 &
  151.9 &
  0.16 &
  4.957 &
  5.199 &
  4.88 &
  12.57 &
  12.81 &
  1.96 &
  10.94 &
  11.18 &
  2.25\\
  \bottomrule
\end{tabular}
}
\end{table}
Table \ref{tab:computational_cost} presents a comprehensive analysis of BSA's computational cost. We measured the model training time (sec/1step), peak memory usage (GB), and the number of parameters (M) for both the base model and the model with BSA applied. The experiments were conducted using the lightweight Weather dataset with 21 channels and the heavy PEMS03 dataset with 358 channels. Tests were performed on Timesnet~\cite{Timesnet}, iTransformer~\cite{iTransformer}, Crossformer~\cite{Crossformer}, and PatchTST~\cite{PatchTST} models. Linear models were excluded from the experiments as they are very lightweight and their training cost is not an issue. Table~\ref{tab:computation} in the main text shows the summarized results.

\subsection{Pre/Post BSA activation signal visualization}
Figure~\ref{fig:spectrogram} demonstrates the transition between pre-activation $F$ (left) and post-activation $F'$ (right) during the inference process. The figure shows the spectrum of randomly sampled activations from the DLinear model trained with a 720 prediction length on the SWDR (W/m) channel of the weather dataset. The x-axis signifies timesteps while the y-axis indicates frequency. To enhance visibility in the low-frequency region, y-axis values below 0.1 are presented in a log scale, whereas values above 0.1 are depicted in a linear scale. A darkening in the high-frequency region is observed from pre-activation to post-activation, suggesting that the activations may undergo a denoising effect in the high-frequency areas as they pass through the SA-Matrix.


\clearpage
\section*{NeurIPS Paper Checklist}

The checklist is designed to encourage best practices for responsible machine learning research, addressing issues of reproducibility, transparency, research ethics, and societal impact. Do not remove the checklist: {\bf The papers not including the checklist will be desk rejected.} The checklist should follow the references and precede the (optional) supplemental material.  The checklist does NOT count toward the page
limit. 

Please read the checklist guidelines carefully for information on how to answer these questions. For each question in the checklist:
\begin{itemize}
    \item You should answer \answerYes{}, \answerNo{}, or \answerNA{}.
    \item \answerNA{} means either that the question is Not Applicable for that particular paper or the relevant information is Not Available.
    \item Please provide a short (1–2 sentence) justification right after your answer (even for NA). 
\end{itemize}

{\bf The checklist answers are an integral part of your paper submission.} They are visible to the reviewers, area chairs, senior area chairs, and ethics reviewers. You will be asked to also include it (after eventual revisions) with the final version of your paper, and its final version will be published with the paper.

The reviewers of your paper will be asked to use the checklist as one of the factors in their evaluation. While "\answerYes{}" is generally preferable to "\answerNo{}", it is perfectly acceptable to answer "\answerNo{}" provided a proper justification is given (e.g., "error bars are not reported because it would be too computationally expensive" or "we were unable to find the license for the dataset we used"). In general, answering "\answerNo{}" or "\answerNA{}" is not grounds for rejection. While the questions are phrased in a binary way, we acknowledge that the true answer is often more nuanced, so please just use your best judgment and write a justification to elaborate. All supporting evidence can appear either in the main paper or the supplemental material provided in the appendix. If you answer \answerYes{} to a question, in the justification please point to the section(s) where related material for the question can be found.

IMPORTANT, please:
\begin{itemize}
    \item {\bf Delete this instruction block, but keep the section heading ``NeurIPS paper checklist"},
    \item  {\bf Keep the checklist subsection headings, questions/answers and guidelines below.}
    \item {\bf Do not modify the questions and only use the provided macros for your answers}.
\end{itemize}


\begin{enumerate}

\item {\bf Claims}
    \item[] Question: Do the main claims made in the abstract and introduction accurately reflect the paper's contributions and scope?
    \item[] Answer:\answerYes{} 
    \item[] Justification: The paper’s contributions and scope are well-explained throughout the abstract and introduction, and the entire paper is written to provide a consistent and comprehensive conclusion.

\item {\bf Limitations}
    \item[] Question: Does the paper discuss the limitations of the work performed by the authors?
    \item[] Answer: \answerYes{} 
    \item[] Justification: Our paper discusses the limitations and future research directions to address them in the conclusion section~\ref{conclusion}.

\item {\bf Theory Assumptions and Proofs}
    \item[] Question: For each theoretical result, does the paper provide the full set of assumptions and a complete (and correct) proof?
    \item[] Answer: \answerNA{} 
    \item[] Justification: Our paper proposes a new model architecture and demonstrates its superiority through a practical approach with extensive experiments rather than theoretical proofs.

    \item {\bf Experimental Result Reproducibility}
    \item[] Question: Does the paper fully disclose all the information needed to reproduce the main experimental results of the paper to the extent that it affects the main claims and/or conclusions of the paper (regardless of whether the code and data are provided or not)?
    \item[] Answer: \answerYes{} 
    \item[] Justification: Our paper provides detailed information to reproduce all experiments. Additionally, the complete experiment code will be provided.

\item {\bf Open access to data and code}
    \item[] Question: Does the paper provide open access to the data and code, with sufficient instructions to faithfully reproduce the main experimental results, as described in supplemental material?
    \item[] Answer: \answerYes{} 
    \item[] Justification: https://github.com/DJLee1208/BSA\_2024

\item {\bf Experimental Setting/Details}
    \item[] Question: Does the paper specify all the training and test details (e.g., data splits, hyperparameters, how they were chosen, type of optimizer, etc.) necessary to understand the results?
    \item[] Answer: \answerYes{} 
    \item[] Justification: Our paper provides detailed information about the experiments, including data splits, hyperparameters, how they were chosen, and the type of optimizer, in the main text and the appendix.

\item {\bf Experiment Statistical Significance}
    \item[] Question: Does the paper report error bars suitably and correctly defined or other appropriate information about the statistical significance of the experiments?
    \item[] Answer: \answerYes{} 
    \item[] Justification: In the main results, we used a paired t-test to demonstrate the superiority of our methodology and reported the p-value.

\item {\bf Experiments Compute Resources}
    \item[] Question: For each experiment, does the paper provide sufficient information on the computer resources (type of compute workers, memory, time of execution) needed to reproduce the experiments?
    \item[] Answer: \answerYes{} 
    \item[] Justification: We reported the types and number of GPUs used in the experiments, as well as the packages used for training.
    
\item {\bf Code Of Ethics}
    \item[] Question: Does the research conducted in the paper conform, in every respect, with the NeurIPS Code of Ethics \url{https://neurips.cc/public/EthicsGuidelines}?
    \item[] Answer: \answerYes{} 
    \item[] Justification: Yes, the research conducted in the paper fully conforms to the NeurIPS Code of Ethics in every respect. We have ensured that all ethical guidelines and standards have been meticulously followed throughout the study.

\item {\bf Broader Impacts}
    \item[] Question: Does the paper discuss both potential positive societal impacts and negative societal impacts of the work performed?
    \item[] Answer: \answerYes{} 
    \item[] Justification: Our paper discusses the potential broader impacts of our research in the conclusion.
    
\item {\bf Safeguards}
    \item[] Question: Does the paper describe safeguards that have been put in place for responsible release of data or models that have a high risk for misuse (e.g., pretrained language models, image generators, or scraped datasets)?
    \item[] Answer: \answerNA{}. 
    \item[] Justification: Our model does not pose such risks, and therefore, safeguards for responsible data or model release were not applicable.

\item {\bf Licenses for existing assets}
    \item[] Question: Are the creators or original owners of assets (e.g., code, data, models), used in the paper, properly credited and are the license and terms of use explicitly mentioned and properly respected?
    \item[] Answer: \answerYes{} 
    \item[] Justification: Yes, we have properly credited the creators or original owners of assets used in the paper, including publicly available models and packages, with accurate citations.

\item {\bf New Assets}
    \item[] Question: Are new assets introduced in the paper well documented and is the documentation provided alongside the assets?
    \item[] Answer: \answerNA{} 
    \item[] Justification: This paper does not include new assets.

\item {\bf Crowdsourcing and Research with Human Subjects}
    \item[] Question: For crowdsourcing experiments and research with human subjects, does the paper include the full text of instructions given to participants and screenshots, if applicable, as well as details about compensation (if any)? 
    \item[] Answer: \answerNA{} 
    \item[] Justification: This paper does not involve crowdsourcing.

\item {\bf Institutional Review Board (IRB) Approvals or Equivalent for Research with Human Subjects}
    \item[] Question: Does the paper describe potential risks incurred by study participants, whether such risks were disclosed to the subjects, and whether Institutional Review Board (IRB) approvals (or an equivalent approval/review based on the requirements of your country or institution) were obtained?
    \item[] Answer:\answerNA{} 
    \item[] Justification: This paper does not involve research with human subjects.

\end{enumerate}

\end{document}